\documentclass{article}

\PassOptionsToPackage{numbers, compress}{natbib}

\usepackage[final]{neurips_2020}

\usepackage[utf8]{inputenc} 
\usepackage[T1]{fontenc}    
\usepackage{hyperref}       
\usepackage{url}            
\usepackage{booktabs}       
\usepackage{amsfonts}       
\usepackage{nicefrac}       
\usepackage{microtype}      

\usepackage{graphicx}
\usepackage{subcaption}
\usepackage{amsmath}
\usepackage{mathtools}
\usepackage{bm}
\usepackage[dvipsnames]{xcolor}
\usepackage[title]{appendix}

\newcommand{\E}{\mathop{\mathbb{E}}}
\newcommand{\Var}{\mathrm{Var}}
\newcommand{\Varpost}{\underset{p(\w_0, \w_1, Y^0,Y^1| \D)}{\Var}}
\DeclareMathOperator{\CATE}{\mathrm{CATE}}
\DeclareMathOperator{\ATE}{\mathrm{ATE}}

\newcommand{\muwone}{\mu^{\w_1}}
\newcommand{\muwzero}{\mu^{\w_0}}
\newcommand{\muzero}{\mu^{0}}
\newcommand{\muone}{\mu^{1}}
\newcommand{\x}{\bm{x}}
\newcommand{\xt}{\x_{\text{test}}}
\newcommand{\w}{\bm{\omega}}
\newcommand{\z}{\bm{z}}
\newcommand{\X}{\bm{X}}

\newcommand{\D}{\mathcal{D}}

\allowdisplaybreaks
\usepackage[small,compact]{titlesec}
\titlespacing{\section}{0pt}{1ex}{0.8ex}
\titlespacing{\subsection}{0pt}{0.5ex}{0ex}
\titlespacing{\subsubsection}{0pt}{0.2ex}{0ex}
\expandafter\def\expandafter\normalsize\expandafter{%
    \normalsize
    \setlength\abovedisplayskip{2pt}
    \setlength\belowdisplayskip{2pt}
    \setlength\abovedisplayshortskip{0pt}
    \setlength\belowdisplayshortskip{0pt}
}

\makeatletter
\newcommand{\printfnsymbol}[1]{%
  \textsuperscript{\@fnsymbol{#1}}%
}
\makeatother

\title{Identifying Causal-Effect Inference Failure with Uncertainty-Aware Models}

\author{%
  Andrew Jesson\thanks{Equal contribution.} \\
  Department of Computer Science\\
  University of Oxford\\
  Oxford, UK OX1 3QD \\
  \texttt{andrew.jesson@cs.ox.ac.uk} \\
  \And
  Sören Mindermann\printfnsymbol{1} \\
  Department of Computer Science\\
  University of Oxford \\
  Oxford, UK OX1 3QD \\
  \texttt{soren.mindermann@cs.ox.ac.uk} \\
  \AND
  Uri Shalit \\
  Technion \\
  Haifa, Israel 3200003 \\
  \texttt{urishalit@technion.ac.il} \\
  \And
  Yarin Gal \\
  Department of Computer Science\\
  University of Oxford \\
  Oxford, UK OX1 3QD \\
  \texttt{yarin.gal@cs.ox.ac.uk} \\
}

\begin{document}

\maketitle

\begin{abstract}
    Recommending the best course of action for an individual is a major application of individual-level causal effect estimation. 
    This application is often needed in safety-critical domains such as healthcare, where estimating and communicating uncertainty to decision-makers is crucial. 
    We introduce a practical approach for integrating uncertainty estimation into a class of state-of-the-art neural network methods used for individual-level causal estimates. 
    We show that our methods enable us to deal gracefully with situations of ``no-overlap'', common in high-dimensional data, where standard applications of causal effect approaches fail. 
    Further, our methods allow us to handle covariate shift, where the train and test distributions differ, common when systems are deployed in practice. 
    We show that when such a covariate shift occurs, correctly modeling uncertainty can keep us from giving overconfident and potentially harmful recommendations. 
    We demonstrate our methodology with a range of state-of-the-art models. 
    Under both covariate shift and lack of overlap, our uncertainty-equipped methods can alert decision makers when predictions are not to be trusted while outperforming standard methods that use the propensity score to identify lack of overlap.
\end{abstract}

\section{Introduction}
\label{sec:introduction}

Learning individual-level causal effects is concerned with learning how units of interest respond to interventions or treatments. 
These could be the medications prescribed to particular patients, training-programs to job seekers, or educational courses for students. 
Ideally, such causal effects would be estimated from randomized controlled trials, but in many cases, such trials are unethical or expensive: researchers cannot randomly prescribe smoking to assess health risks. 
Observational data offers an alternative, with typically larger sample sizes and lower costs, and more relevance to the target population. 
However, the price we pay for using observational data is lower certainty in our causal estimates, due to the possibility of unmeasured confounding, and the measured and unmeasured differences between the populations who were subject to different treatments.

Progress in learning individual-level causal effects is being accelerated by deep learning approaches to causal inference \citep{johansson2016learning,cevae_louizos2017causal,atan2018deep,dragon2019}. 
Such neural networks can be used to learn causal effects from observational data, but current deep learning tools for causal inference cannot yet indicate when they are unfamiliar with a given data point. 
For example, a system may offer a patient a recommendation even though it may not have learned from data belonging to anyone with similar age or gender as the patient, or it may have never observed someone like this patient receive a specific treatment before. 
In the language of machine learning and causal inference, the first example corresponds to a \emph{covariate shift}, and the second example corresponds to a violation of the \emph{overlap assumption}, also known as positivity. 
When a system experiences either covariate shift or violations of overlap, the recommendation would be uninformed and could lead to undue stress, financial burden, false hope, or worse. 
In this paper, we explain how covariate shift and violations of overlap are concerns for real-world learning of conditional average treatment effects (CATE) from observational data, we examine why deep learning systems should indicate their lack of confidence when these phenomena are encountered, and we develop a new and principled approach to incorporating uncertainty estimating into the design of systems for CATE inference.

First, we reformulate the lack of overlap at test time as an instance of covariate shift, allowing us to address both problems with one methodology. 
When an observation $\x$ lacks overlap, the model predicts the outcome $y$ for a treatment $t$ that has probability zero or near-zero under the training distribution. 
We extend the Causal-Effect Variational Autoencoder (CEVAE) \citep{cevae_louizos2017causal} by incorporating negative sampling, a method for out-of-distribution (OoD) training, to improve uncertainty estimates on OoD inputs. 
Negative sampling is effective and theoretically justified but usually intractable \citep{hafner2018reliable}. 
Our insight is that it becomes tractable for addressing non-overlap since the distribution of test-time inputs $(\x,t)$ is known: it equals the training distribution but with a different choice of treatment (for example, if at training we observe outcome $y$ for patient $\x$ only under treatment $t=0$, then we know that the outcome for $(\x, t=1)$ should be uncertain). 
This can be seen as a special case of transductive learning \citep[Ch. 9]{vapnik1999nature}. 
For addressing covariate shift in the inputs $\x$, negative sampling remains intractable as the new covariate distribution is unknown; however, it has been shown in non-causal applications that Bayesian parameter uncertainty captures ``epistemic'' uncertainty which can indicate covariate shift \citep{kendall2017uncertainties}. 
We, therefore, propose to treat the decoder $p(y|\x, t)$ in CEVAE as a Bayesian neural network able to capture epistemic uncertainty. 

In addition to casting lack of overlap as a distribution shift problem and proposing an OoD training methodology for the CEVAE model, we further extend the modeling of epistemic uncertainty to a range of state-of-the-art neural models including TARNet, CFRNet \citep{shalit2017estimating}, and Dragonnet \citep{shi2019adapting}, developing a practical Bayesian counter-part to each. 
We demonstrate that, by excluding test points with high epistemic uncertainty at test time, we outperform baselines that use the propensity score $p(t=1|\x)$ to exclude points that violate overlap. 
This result holds across different state-of-the-art architectures on the causal inference benchmarks IHDP \citep{hill2011bayesian} and ACIC \citep{dorie2019automated}. 
Leveraging uncertainty for exclusion ties it into causal inference practice where a large number of overlap-violating points must often be discarded or submitted for further scrutiny \citep{rosenbaum1983central, imbens2004nonparametric, crump2009dealing, imbens2015causal, hernan2010causal}. 
Finally, we introduce a new semi-synthetic benchmark dataset, CEMNIST, to explore the problem of non-overlap in high-dimensional settings.

\begin{figure*}[ht!]
    \centering
    \begin{subfigure}[]{\textwidth}
        \centering
        \includegraphics[width=\textwidth]{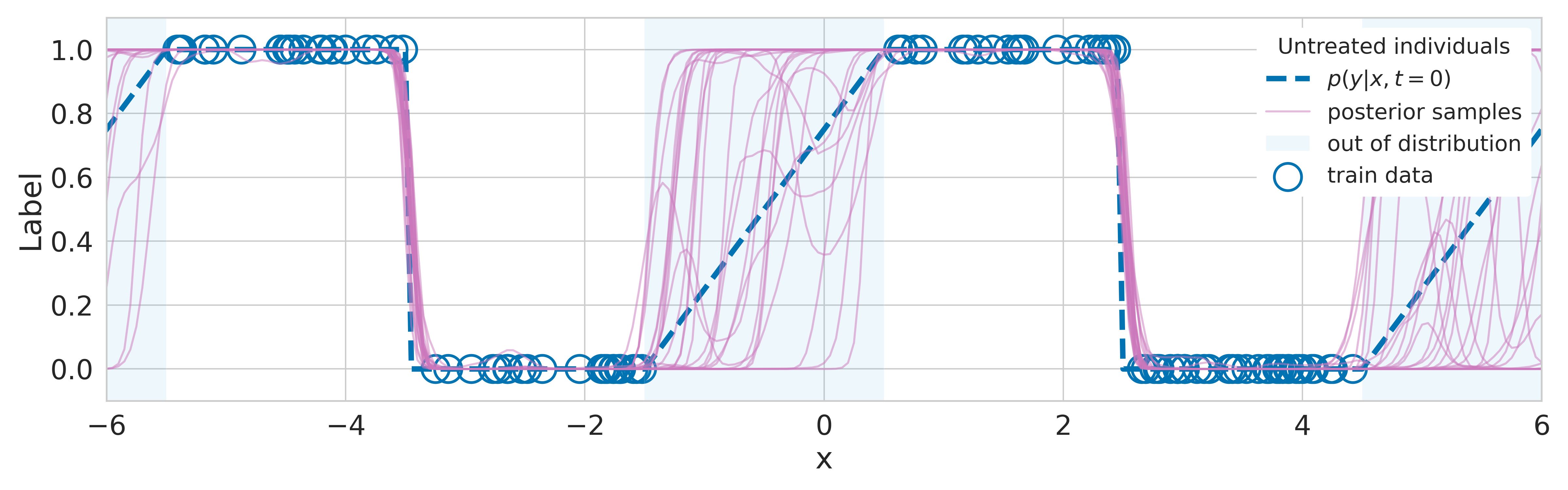}
        \label{fig:intuitive_model_0}
    \end{subfigure}
    ~ \vspace{-12mm}
    
    \begin{subfigure}[]{\textwidth}
        \centering
        \includegraphics[width=\textwidth]{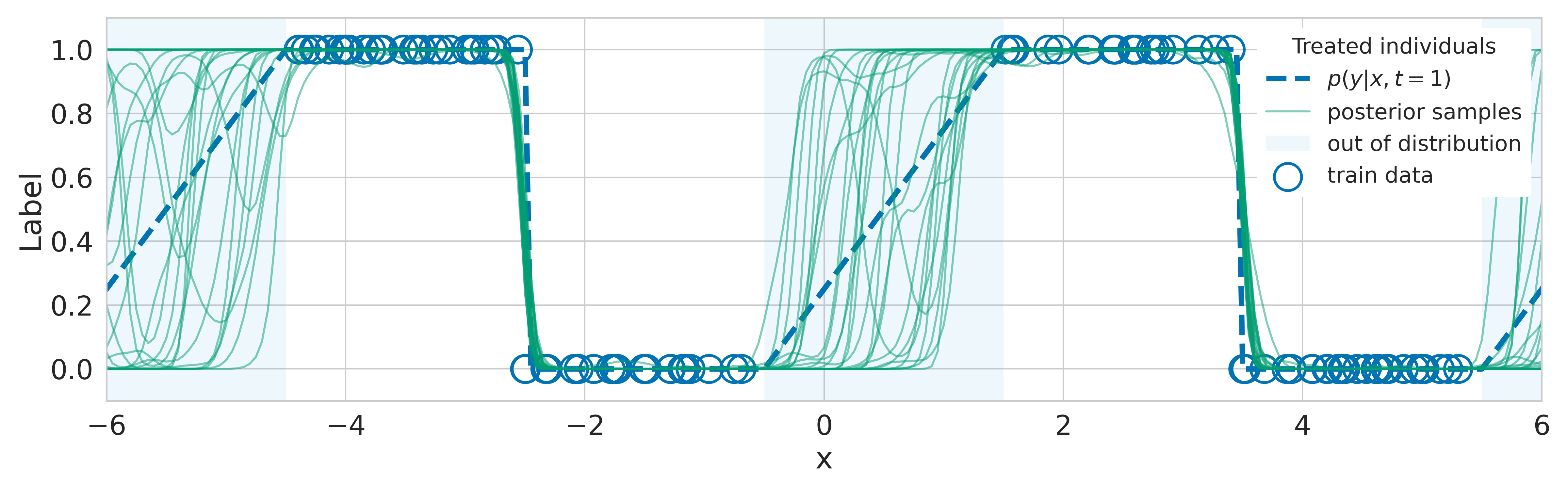}
        \label{fig:intuitive_model_1}
    \end{subfigure}
    ~ \vspace{-12mm}
    
    \begin{subfigure}[]{\textwidth}
        \centering
        \includegraphics[width=\textwidth]{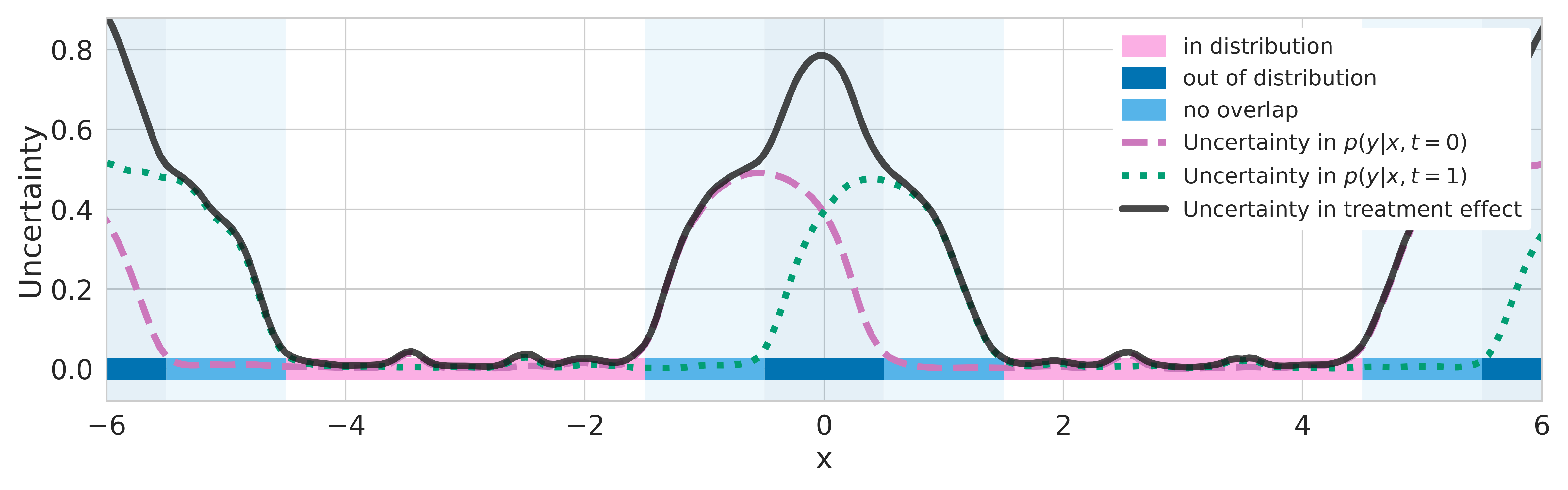}
        \label{fig:intuitive_epistemic_uncertainty}
    \end{subfigure}
    \vspace{-6mm}
    \caption{
        Explanation of how epistemic uncertainty detects lack of data. 
        \textbf{Top}: binary outcome $y$ (blue circle) given \textit{no} treatment, and different functions $p(y=1|\x,t=0, \w)$ (purple) predicting outcome probability (blue dashed line, ground truth). 
        Functions disagree where data is scarce. 
        \textbf{middle}: binary outcome $y$ given treatment, and functions $p(y=1|\x,t=1, \w)$ (green) predicting outcome  probability. 
        \textbf{Bottom}: measures of uncertainty/disagreement between outcome predictions (dashed purple and dotted green lines) are high when data is lacking. CATE uncertainty (solid black line) is higher where at least one model lacks data (non-overlap, light blue) or where both lack data (out-of-distribution / covariate shift, dark blue).
    }
    \label{fig:intuitive_uncertainty}
    \vspace{-4mm}
\end{figure*}

\section{Background}
\label{sec:background}

Classic machine learning is concerned with functions that map an input (e.g. an image) to an output (e.g. ``is a person''). 
The specific function $f$ for a given task is typically chosen by an algorithm that minimizes a loss between the outputs $f(\x_i)$ and targets $y_i$ over a dataset $\{\x_i, y_i\}_{i=1}^{N}$ of input covariates and output targets. 
Causal effect estimation differs in that, for each input $\x_i$,  there is a corresponding treatment $t_i\in\{0,1\}$ and two potential outcomes $Y^1$, $Y^0$ -- one for each choice of treatment \citep{rubin2005causal}. 
In this work, we are interested in the Conditional Average Treatment Effect ($\CATE$):
\begin{align}
    \CATE(\x_i) &= \E[Y^1-Y^0|\X=\x_i] \\
    &= \muone(\x_i) - \muzero(\x_i),
    \label{CATE}
\end{align}
where the expectation is needed both because the \textit{individual treatment effect} $Y^1-Y^0$ may be non-deterministic, and because it cannot in general be identified without further assumptions. 
Under the assumption of ignorability conditioned on $\X$ (or no-hidden confounding), which we make in this paper, we have that $\E[Y^a|\X=\x_i] = \E[y|\X=\x_i,t=a]$, thus opening the way to estimate CATE from observational data \citep{imbens2015causal}. 
Specifically, we are motivated by cases where $\X$ is high-dimensional, for example, a patient's entire medical record, in which case we can think of the CATE as representing an individual-level causal effect. 
Though the specific meaning of a CATE measurement depends on context, in general, a positive value indicates that an individual with covariates $\x_i$ will have a positive response to treatment, a negative value indicates a negative response, and a value of zero indicates that the treatment will not affect such an individual.

The fundamental problem of learning to infer $\CATE$ from an observational dataset $\mathcal{D} = \{\x_i, y_i, t_i\}_{i=1}^{N}$ is that only the \textit{factual} outcome $y_i=Y^{t_i}$ corresponding to the treatment $t_i$ can be observed. 
Because the \textit{counterfactual} outcome $Y^{1-t_i}$ is never observed, it is difficult to learn a function for $\CATE(\x_i)$ directly. 
Instead, a standard approach are either to treat $t_i$ as an additional covariate \citep{gelman2007causal} or focus on learning functions for $\muzero(\x_i)$ and $\muone(\x_i)$ using the observed $y_i$ in $\mathcal{D}$ as targets \citep{shalit2017estimating, cevae_louizos2017causal, dragon2019}.

\subsection{Epistemic uncertainty and covariate shift}

In probabilistic modelling, predictions may be assumed to come from a graphical model $p(y | \x, t, \w)$ -- a distribution over outputs (the likelihood) given a single set of parameters $\w$.
Considering a binary label $y$ given, for example, $t=0$, a neural network can be described as a function defining the likelihood $p(y=1|\x, t=0, \w_0)$, with parameters $\w_0$ defining the network weights. 
Different draws $\w_0$ from a distribution over parameters $p(\w_0 | \D)$ would then correspond to different neural networks, i.e. functions from $(\x, t=0)$ to $y$ (e.g. the purple curves in Fig. \ref{fig:intuitive_uncertainty} (top)).

For parametric models such as neural networks (NNs), we treat the weights as random variables, and, with a chosen prior distribution $p(\w_0)$, aim to infer the posterior distribution  $p(\w_0 | \D)$. 
The purple curves in Figure \ref{fig:intuitive_uncertainty} (top) are individual NNs $\mu^{\w_0}(\cdot)$ sampled from the posterior of such a Bayesian neural network (BNN). 
\textit{Bayesian inference} can be performed by marginalizing the likelihood function $p(y | \mu^{\w_0}(\x))$ over the posterior $p(\w_0|\D)$ in order to obtain the posterior predictive probability $p(y|\x, t=0, \mathcal{D}) = \int p(y | \x, t=0, \w_0) p(\w_0 | \mathcal{D}) d\w_0$. 
This marginalization is intractable for BNNs in practice, so variational inference is commonly used as a scalable approximate inference technique, for example, by sampling the weights from a Dropout approximate posterior $q(\w_0|\mathcal{D})$ ~\citep{gal2016dropout}.

Figure \ref{fig:intuitive_uncertainty} (top) illustrates the effects of a BNN's parameter uncertainty in the range $\x \in [-1, 1]$ (shaded region). 
While all sampled functions $\mu^{\w_0}(\x)$ with $\w_0 \sim p(\w_0 | \D, t=0)$ (shown in blue) agree with each other for inputs $\x$ in-distribution ($\x \in [-6, -1]$) these functions make disagreeing predictions for inputs $\x \in [-1, 1]$ because these lie out-of-distribution (OoD) with respect to the training distribution $p(\x | t = 0)$. This is an example of \emph{covariate shift}.

To avoid overconﬁdent erroneous extrapolations on such OoD examples, we would like to indicate that the prediction $\mu^{\w_0}(\x)$ is uncertain. 
This \emph{epistemic} uncertainty stems from a lack of data, and is distinct from \emph{aleatoric} uncertainty, which stems from measurement noise. 
Epistemic uncertainty about the random variable (r.v.) $Y^0$ can be quantified in various ways. 
For classification tasks, a popular information-theoretic measure is the information gained about the r.v. $\w_0$ if the label $y=Y^0$ were observed for a new data point $\x$, given the training dataset $\D$ \citep{houlsby2011bayesian}. 
This is captured by the mutual information between $\w_0$ and $Y^0$, given by

\begin{equation}
    \mathcal{I}[\w_0, Y^0 | \D, \x] = \mathcal{H}[Y^0 | \x, \D] \quad - \E_{q(\w_0 | \D)} \left[\mathcal{H}[Y^0 | \x, \w_0]\right],
\end{equation}

where $\mathcal{H}[\bullet]$ is the entropy of a given r.v.. 
For regression tasks, it is common to measure how the r.v. $\muwzero(\x)$ varies when marginalizing over $\w_0$: $\underset{_{q(\w_0|\D)}}{\mathrm{Var}}[\muwzero ( \x)]$.
We will later use this measure for epistemic uncertainty in $\CATE$.

\section{Non-overlap as a covariate shift problem}
\label{sec:distribution_shift}

Standard causal inference tasks, under the assumption of ignorability conditioned on $\X$, usually deal with estimating both $\mu^0(\x)~=~\E[y|\X=\x,t=0]$ and $\mu^1(\x)~=~\E[y|\X=\x,t=1]$. 
Overlap is usually assumed as a means to address this problem. 
The overlap assumption (also known as \textit{common support} or \textit{positivity}) states that there exists $0<\eta<0.5$ such that the \textit{propensity score} $p(t=1|\x)$ satisfies:
\begin{equation}  \label{eq:overlap}
    \eta<p(t=1|\x)<1-\eta,  \quad
\end{equation}
i.e., that for every $\x\sim p(\x)$, we have a non-zero probability of observing its outcome under $t=1$ as well as under $t=0$. 
This version is sometimes called \emph{strict overlap}, see \citep{d2017overlap} for discussion. 
When overlap does not hold for some $\x$, we might lack data to estimate either $\mu^0(\x)$ or $\mu^1(\x)$---this is the case in the grey shaded areas in Figure \ref{fig:intuitive_uncertainty} (bottom). 

Overlap is a central assumption in causal inference \citep{rosenbaum1983central, imbens2004nonparametric}. 
Nonetheless, it is usually not satisfied for all units in a given observational dataset \citep{rosenbaum1983central, imbens2004nonparametric, crump2009dealing, imbens2015causal, hernan2010causal}. 
It is even harder to satisfy for high-dimensional data such as images and comprehensive demographic data \citep{d2017overlap} where neural networks are used in practice \citep{goodfellow2016deep}. 

Since overlap must be assumed for most causal inference methods, an enormously popular practice is ``trimming'': removing the data points for which overlap is not satisfied before training \citep{hernan2010causal, fogarty2016discrete, dragon2019, king2019propensity, d1998propensity}. 
In practice, points are trimmed when they have a propensity close to 0 or 1, as predicted by a trained propensity model $p^{\w_p}(t | \x)$. 
The average treatment effect (ATE), is then calculated by over the remaining training points. 

However, trimming has a different implication when estimating the CATE for each unit with covariates $\x_i$: it means that for some units a CATE estimate is not given. 
If we think of CATE as a tool for recommending treatment assignment, a trimmed unit receives no treatment recommendation. 
This reflects the uncertainty in estimating one of the potential outcomes for this unit, since treatment was rarely (if ever) given to similar units. 
In what follows, we will explore how trimming can be replaced with more data-efficient rejection methods that are specifically focused on assessing the level of uncertainty in estimating the expected outcomes for $\x_i$ under both treatment options. 

Our model of the $\CATE$ is:
\begin{equation}
    \widehat{\CATE}^{\w_{0/1}}(\x) =~\muwone(\x) - \muwzero(\x).
    \label{eq:cate}
\end{equation}
In Figure \ref{fig:intuitive_uncertainty}, we illustrate that lack of overlap constitutes a covariate shift problem. 
When $p(t=1|\xt)\approx 0$, we face a covariate shift for $\muwone(\cdot)$  (because by Bayes rule $p(\xt | t=1)\approx 0$). 
When $p(t=1 | \xt)\approx 1$, we face a covariate shift for $\muwzero(\cdot)$, and when $p(\xt)\approx 0$, we face a covariate shift for $\widehat{\CATE}^{\w_{0/1}}(\x)$ (``out of distribution'' in the Figure~\ref{fig:intuitive_uncertainty} (bottom)). 
With this understanding, we can deploy tools for epistemic uncertainty to address both covariate shift and non-overlap simultaneously.

\subsection{Epistemic uncertainty in CATE}
To the best of our knowledge, uncertainty in high-dimensional $\CATE$ (i.e. where each value of $\x$ is only expected to be observed at most once) has not been previously addressed. 
$\CATE(\x)$ can be seen as the first moment of the random variable $Y^1-Y^0$ given $\X=\x$. 
Here, we extend this notion and examine the \textit{second} moment, the variance, which we can decompose into its aleatoric and epistemic parts by using the law of total variance:

\begin{equation}
    \begin{split}
        \Varpost[Y^1 - Y^0 | \x] 
        & = \E_{p(\w_0, \w_1| \D)}\left[\underset{Y_0,Y_1}{\Var} [Y^1-Y^0 ~~|~~ \muwone(\x),\muwzero(\x)]\right] \\
        & ~+~ \underset{p(\w_0,\w_1|\D)}{\Var}[\muwone(\x) - \muwzero(\x)].
    \end{split}
\label{eq:variance_in_cate}
\end{equation}

The second term on the r.h.s. is $\Var[\widehat{\CATE}^{\w_{0/1}}(\x)]$. 
It measures the epistemic uncertainty in $\CATE$ since it only stems from the disagreement between predictions for different values of the parameters, not from noise in $Y^1,Y^0$. 
We will use this uncertainty in our methods and estimate it directly by sampling from the approximate posterior $q(\w_0, \w_1 | \D)$. 
The first term on the r.h.s. is the expected aleatoric uncertainty, which is disregarded in $\CATE$ estimation (but could be relevant elsewhere).

Referring back to Figure \ref{fig:intuitive_uncertainty}, when overlap is not satisfied for $\x$, $\Var[\widehat{\CATE}^{\w_{0/1}}(\x)]$ is large because at least one of $\Var_{\w_0}[\muwzero(\x)]$ and $\Var_{\w_1}[\muwone(\x)]$ is large. 
Similarly, under regular covariate shift ($p(\x)\approx 0$), both will be large.

\subsection{Rejection policies with epistemic uncertainty versus propensity score} \label{sec:rej_policies}

If there is insufficient knowledge about an individual, and a high cost associated with making errors, it may be preferable to withhold the treatment recommendation. 
It is therefore important to have an informed \emph{rejection policy}. 
In our experiments, we reject, i.e. choose to make no treatment recommendation, when the epistemic uncertainty exceeds a certain threshold. 
In general, setting the threshold will be a domain-specific problem that depends on the cost of type I (incorrectly recommending treatment) and type II (incorrectly withholding treatment) errors. 
In the diagnostic setting, thresholds have been set to satisfy public health authority specifications, e.g. for diabetic retinopathy \cite{leibig2017leveraging}. 
Some rejection methods additionally weigh the chance of algorithmic error against that of human error \cite{raghu2019algorithmic}.

When instead using the propensity score for rejection, a simple policy is to specify $\eta_0$ and reject points that do not satisfy eq. \eqref{eq:overlap} with $\eta = \eta_0$. 
More sophisticated standard guidelines were proposed by \citet{caliendo2008some}. 
These methods only account for the uncertainty about $\CATE(\x)$ that is due to limited overlap and do not consider that uncertainty is also modulated by the availability of data on similar individuals (as well as the noise in this data).

\section{Adapting neural causal models for covariate shift}
\label{sec:methods}

\subsection{Parameter uncertainty}
\label{sec:bayes_decod}
To obtain the epistemic uncertainty in the $\CATE$, we must infer the parameter uncertainty distribution conditioned on the training data $p(\w_0,\w_1|\D)$, which defines the distribution of each network $\muwzero(\cdot),\muwone(\cdot)$, conditioned on $\D$. 
There exists a large suite of methods we can leverage for this task, surveyed in \citet{gal2016uncertainty}. 
Here, we use MC Dropout \citep{gal2016dropout} because of its high scalability \citep{tran2019bayesian}, ease of implementation, and state-of-the-art performance \citep{filos2019systematic, zhu2017deep, mcallister2017concrete, jungo2017towards}. 
However, our contributions are compatible with other approximate inference methods. 
\citet{gal2016dropout} has shown that we can simply add dropout \citep{srivastava2014dropout} with L2 regularization in each of $\w_0,\w_1$ during training and then sample from the same dropout distribution at test time to get samples from $q(\w_0,\w_1|\D)$. 
With tuning of the dropout probability, this is equivalent to sampling from a Bernoulli approximate posterior $q(\w_0,\w_1|\D)$ (with standard Gaussian prior). 
While most neural causal inference methods can be adapted in this way, CEVAE \citep{cevae_louizos2017causal} is more complicated and will be addressed in the next section.

\subsection{Bayesian CEVAE}
\label{sec:cevae}

The Causal Effect Variational Autoencoder (CEVAE, \citet{cevae_louizos2017causal}) was introduced as a means to relax the common assumption that the data points $\x_i$ contain accurate measurements of all confounders -- instead, it assumes that the observed $\x_i$ are a noisy transformation of some true confounders $\z_i$, whose conditional distribution can nonetheless be recovered. 
To do so, CEVAE encodes each observation $(\x_i,t_i,y_i)\in \D$, into a distribution over latent confounders $\z_i$ and reconstructs the entire observation with a decoder network. 
For each possible value of $t\in\{0,1\}$, there is a separate branch of the model. 
For each branch $j$, the encoder has an auxiliary distribution $q(y_i|\x_i,t=j)$ to approximate the posterior $q(\mathbf{z}_i|\mathbf{x}_i,y_i,t=j)$ at test time. 
It additionally has a single auxiliary distribution $q(t_i|\mathbf{x}_i)$ which generates $t_i$. 
See Figure 2 in \cite{cevae_louizos2017causal} for an illustration. 
The decoder reconstructs the entire observation, so it learns the three components of $p(\mathbf{x}_i,t_i,y_i|\mathbf{z}_i)~=~p(t_i|\mathbf{z}_i)~p(y_i|t_i,\mathbf{z}_i)~p(\mathbf{x}_i|\mathbf{z}_i)$. 
We will omit the parameters of these distributions to ease our notation. 
The encoder parameters are summarized as $\psi$ and the decoder parameters as $\w$.

If the treatment and outcome were known at test time, the training objective (ELBO) would be
\begin{equation}
    \mathcal{L}  = \sum_{i=1}^{N} \E_{q\left(\mathbf{z}_{i} | \mathbf{x}_{i}, t_{i}, y_{i}\right)} \big[\log p\left(\mathbf{x}_{i}, t_{i} |\mathbf{z}_{i}\right)
    + \log p\left(y_{i} | t_{i}, \mathbf{z}_{i}\right)\big] - KL(q(\mathbf{z}_{i} | \mathbf{x}_{i}, t_{i}, y_{i})  \ || \ p(\mathbf{z}_{i}))
    \label{ELBO_VAE}
\end{equation}

where $KL$ is the Kullback-Leibler divergence. 
However, $t_i$ and $y_i$ need to be predicted at test time, so CEVAE learns the two additional distributions by using the objective

\begin{equation}     \label{ELBO_CEVAE}
    \mathcal{F} = \mathcal{L} + \sum_{i=1}^N (\log q(t_i=t_i^*|\mathbf{x}_i) + \log q(y_i=y_i^*|\mathbf{x}_i,t_i^*)),
\end{equation}

where a star indicates that the variable is only observed at training time. 
At test time, we calculate the $\CATE$ so $t_i$ is set to $0$ and $1$ for the corresponding branch and $y_i$ is sampled both times.

Although the encoder performs Bayesian inference to infer $\z_i$, CEVAE does not model epistemic uncertainty because the decoder lacks a distribution over $\w$. 
The recently introduced Bayesian Variational Autoencoder \citep{daxberger2019bayesian} attempts to model such epistemic uncertainty in VAEs using MCMC sampling. 
We adapt their model for causal inference by inferring an approximate posterior $q(\w|\mathcal{D})$. 
In practice, this is again a simple change if we use Monte Carlo (MC) Dropout in the decoder
\footnote{
    We do not treat the parameters $\psi$ of the encoder distributions as random variables. 
    This is because the encoder does not infer $\z$ directly. 
    Instead, it parameterizes the \textit{parameters} $\mu(\z), \Sigma(\z)$ of a Gaussian posterior over $\z$ (see eq. (5) in \citet{cevae_louizos2017causal} for details). These parameters specify the uncertainty over $\z$ themselves.
}. 
This is implemented by adding dropout layers to the decoder and adding a term $KL(q(\w | \D) || p(\w))$ to eq. \eqref{ELBO_CEVAE}, where $p(\w)$ is standard  Gaussian. 
Furthermore, the expectation in eq. \eqref{ELBO_VAE} now goes over the \textit{joint} posterior $q(\mathbf{z}_{i} | \mathbf{x}_{i}, t_{i}, y_{i})q(\w|\D)$ by performing stochastic forward passes with Dropout `turned on'. Likewise, the joint posterior is used in the right term of eq. \eqref{eq:variance_in_cate}.

\textbf{Negative sampling for non-overlap.} 
\textit{Negative sampling} is a powerful method for modeling uncertainty under a covariate shift by adding loss terms that penalize confident predictions on inputs sampled outside the training distribution \citep{sun2019functional, lee2017training, hafner2018reliable, hendrycks2018deep, rowley1998neural}. 
However, it is usually intractable because the $\x$ input space is high dimensional. 
Our insight is that it becomes tractable for non-overlap, because the OoD inputs are created by simply flipping $t$ on the in-distribution inputs $\{(\x_i,t_i)\}$ to create the new inputs $\{(\x_i,t_i'=1-t_i)\}$. 
Our negative sampling is implemented by mapping each $(\x_i,y_i,t_i)\in\D$ through \textit{both} branches of the encoder. 
On the \textit{counterfactual} branch, where $t_i'=1-t_i$, we only minimize the KL divergence from the posterior $q(\z|\xt,t=0)$ to $p(\z)$, but none of the other terms in eq. \eqref{ELBO_CEVAE}. 
This is to encode that we have no information on the counterfactual prediction. 
In appendix \ref{app:results:negative_sampling} we study negative sampling and demonstrate improved uncertainty.

\section{Related work}
\label{sec:related_works}
Epistemic uncertainty is modeled out-of-the-box by non-parametric Bayesian methods such as Gaussian Processes (GPs) \citep{rasmussen2003gaussian} and Bayesian Additive Regression Trees (BART) \cite{chipman2010bart}. 
Various non-parametric models have been applied to causal inference \citep{alaa2017bayesian,chipman2010bart,zhang2020learning, hill2011bayesian, wager2018estimation}. 
However, recent state-of-the-art results for high-dimensional data have been dominated by neural network approaches \citep{johansson2016learning,cevae_louizos2017causal,atan2018deep,dragon2019}. 
Since these do not incorporate epistemic uncertainty out-of-the-box, our extensions are meant to fill this gap in the literature.

Causal effects are usually estimated after discarding/rejecting points that violate overlap, using the estimated propensity score \citep{crump2009dealing, hernan2010causal, fogarty2016discrete, dragon2019, king2019propensity, d1998propensity}. 
This process is cumbersome, and results are often sensitive to a large number of ad hoc choices \citep{hill2011challenges} which can be avoided with our methods.  
\citet{hill2013assessing} proposed alternative heuristics for discarding by using the epistemic uncertainty provided by BART on low dimensional data, but focuses on learning the $\ATE$, the average treatment effect over the training set, and neither uses uncertainty in $\CATE$ nor $\ATE$.

In addition to violations of overlap, we also address $\CATE$ estimation for \textit{test} data. 
Test data introduces the possibility of covariate shift away from $p(\x)$, which has been studied outside the causal inference literature \citep{quionero2009dataset,li2011knows,sugiyama2007covariate,shimodaira2000improving}. 
In both cases, we may wish to reject $\x$, e.g. to consult a human expert instead of making a possibly false treatment recommendation. 
To our knowledge, there has been no comparison of rejection methods for $\CATE$ inference.

\section{Experiments}
\label{sec:experiments}

In this section, we show empirical evidence for the following claims: 
that our uncertainty aware methods are robust both to violations of the overlap assumption and a failure mode of propensity based trimming (\ref{sec:experiment_overlap}); 
that they indicate high uncertainty when covariate shifts occur between training and test distributions (\ref{sec:experiment_cov_shift}); 
and that they yield lower $\CATE$ estimation errors while rejecting fewer points than propensity based trimming (\ref{sec:experiment_cov_shift}). 
In the process, we introduce a new, high-dimensional, individual-level causal effect prediction benchmark dataset called CEMNIST to demonstrate robustness to overlap and propensity failure (\ref{sec:experiment_overlap}). 
Finally, we introduce a modification to the IHDP causal inference benchmark to explore covariate shift (\ref{sec:experiment_cov_shift}).

We evaluate our methods by considering \emph{treatment recommendations}. 
A simple treatment recommendation strategy assigns $t=1$ if the predicted $\widehat{\CATE}(x_i)$ is positive, and $t=0$ if negative.
As stated in section \ref{sec:rej_policies}, insufficient knowledge about an individual and high costs due to error necessitate informed \emph{rejection policies} to formalize when a recommendation should be withheld. 
We compare four rejection policies: \emph{epistemic uncertainty} using $\Var[\widehat{\CATE}^{\w_{0/1}}(\x)]$, \emph{propensity quantiles}, \emph{propensity trimming} \citep{caliendo2008some} and \emph{random} (implementation details of each policy are given in Appendix \ref{app:evaluation}). 
Policies are ranked according to the proportion of incorrect recommendations made, given a fixed rate ($r_\text{rej}$) of withheld recommendations. 
This corresponds to assigning a cost of 1 to making an incorrect prediction and a cost of 0 for either making a correct recommendation or withholding an automated recommendation and deferring the decision to a human expert instead. 
We also report the Precision in Estimation of Heterogeneous Treatment Effect (PEHE) \citep{hill2011bayesian, shalit2017estimating} over the non-rejected subset. 
The mean and standard error of each metric is reported over a dataset-dependent number of training runs.

We evaluate and compare each rejection policy using several uncertainty aware $\CATE$ estimators. 
The estimators are Bayesian versions of CEVAE \cite{cevae_louizos2017causal}, TARNet, CFR-MMD \citep{shalit2017estimating}, Dragonnet \citep{dragon2019}, and a deep T-Learner. 
Each model is augmented by introducing Bayesian parameter uncertainty and by predicting a distribution over model outputs. 
For imaging experiments, a two-layer CNN encoder is added to each model. 
Details for each model are given in Appendix \ref{app:models}. 
In the result tables, each model's name is prefixed with a ``B" for ``Bayesian''. 
We also compare to Bayesian Additive Regression Trees (BART) \cite{hill2011bayesian}.

\subsection{Using uncertainty when overlap is violated}
\label{sec:experiment_overlap}

\textbf{Causal effect MNIST (CEMNIST).} We introduce the CEMNIST dataset using hand-written digits from the MNIST dataset \citep{lecun1998mnist} to demonstrate that our uncertainty measures capture non-overlap on high-dimensional data and that they are robust to a failure mode of propensity score rejection. 

\begin{table}[h!] \centering
    \vspace{-5mm}
    \caption{\textbf{CEMNIST-Overlap} Description of ``Causal effect MNIST'' dataset.}
    \vspace{0.1in}
    \begin{small}
        \begin{tabular}{@{}lccccc@{}}\toprule
            \textbf{Digit(s)} & $p(\x)$ & $p(t=1 | \x)$ & $p(y=1 | \x, t=0)$ & $p(y=1 | \x, t=1)$ & $\CATE$   \\ \midrule
            9 & $0.5$\         & $1 / 9$\ & $1$\  & $0$\  & $-1$\  \\ 
            2 & $0.5 / 9$\ & $1$\          & $0$\  & $1$\  & $1$\   \\ 
            other odds & $0.5 / 9$\ & $0.5$\        & $1$\  & $0$\  & $-1$\  \\ 
            other evens & $0.5 / 9$\ & $0.5$\        & $0$\  & $1$\  & $1$\  \\
            \bottomrule
        \end{tabular}
    \end{small}
    \label{tab:cemnist-overlap}
\end{table}

Table \ref{tab:cemnist-overlap} depicts the data generating process for CEMNIST. 
In expectation, half of the samples in a generated dataset will be nines, and even though the propensity for treating a nine is relatively low, there are still on average twice as many treated nines as there are samples of other treated digits (except for twos). 
Therefore, it is reasonable to expect that the $\CATE$ can be estimated most accurately for nines. 
For twos, there is strict non-overlap. 
Therefore, the $\CATE$ cannot be estimated accurately. 
For the remaining digits, the $\CATE$ estimate should be less confident than for nines because there are fewer examples during training, but more confident than for twos because there are both treated and untreated training examples.

\begin{figure*}[h!]
    \centering
    \hspace{-5mm}
    \begin{subfigure}[t]{0.30\textwidth}
        \centering
        \includegraphics[width=\textwidth]{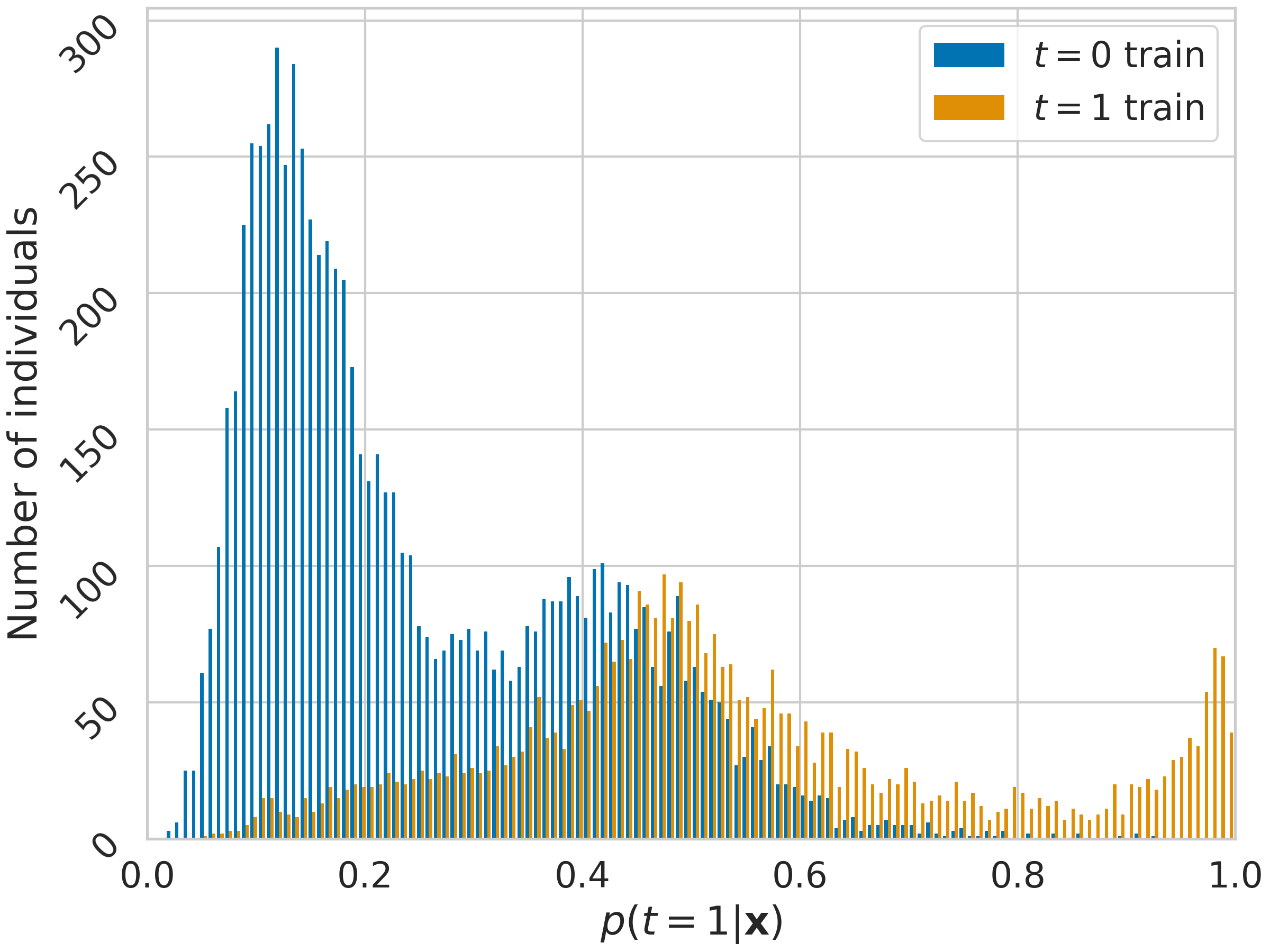}
        \caption{Propensity histogram}
        \label{fig:cemnist_vis_hist}
    \end{subfigure} \hspace{-4mm}
    ~ 
    \begin{subfigure}[t]{0.30\textwidth}
        \centering
        \includegraphics[width=\textwidth]{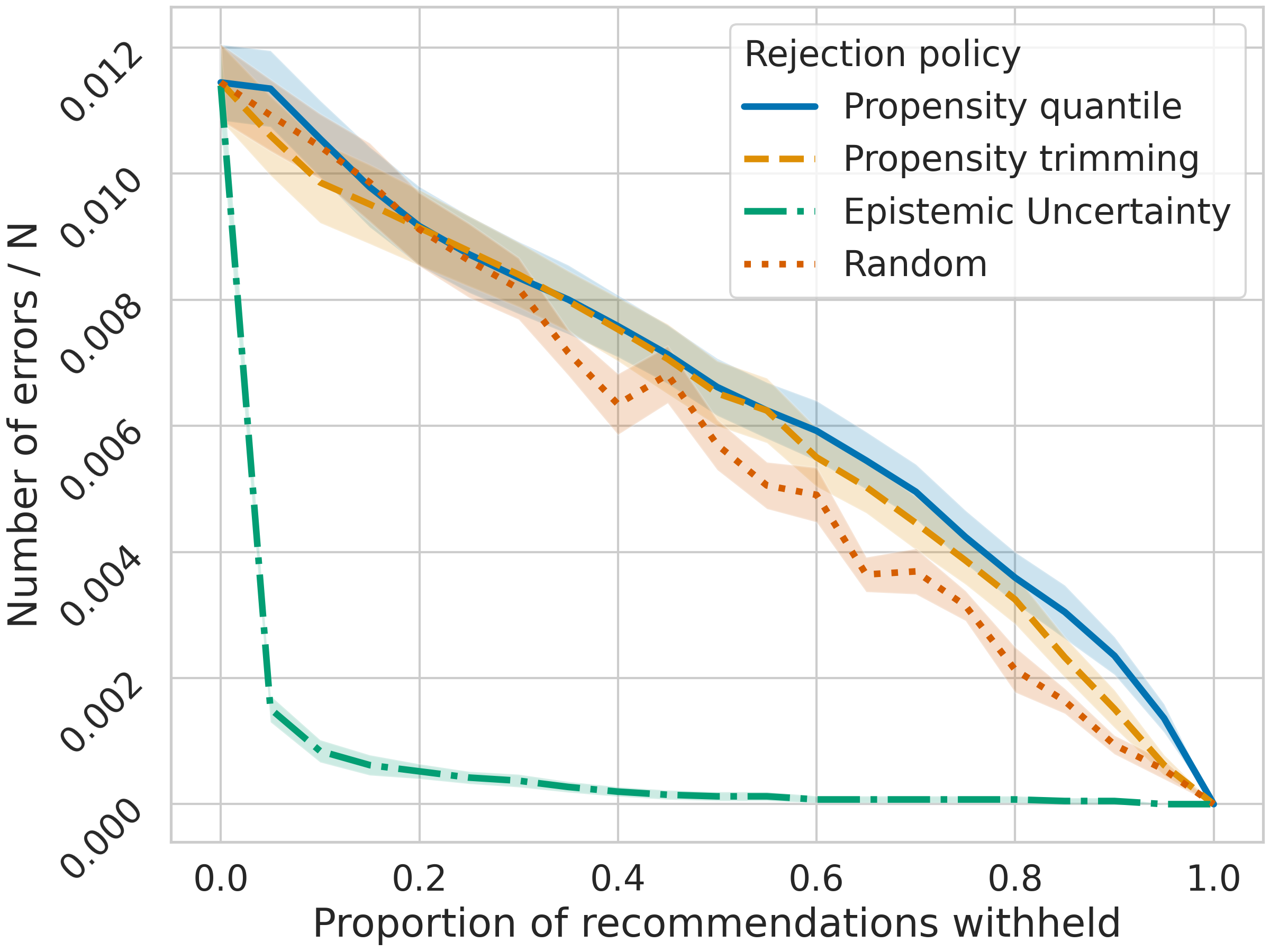}
        \caption{Error rate vs. $r_\text{rej}$}
        \label{fig:cemnist_vis_error}
    \end{subfigure} \hspace{-5mm}
    ~ 
    \begin{subfigure}[t]{0.40\textwidth} 
        \centering
        \subfloat[t][Model comparison\label{fig:cemnist_table}]{
            \begin{small}
                \begin{tabular}{@{}lrrr@{}} \toprule
                    \textbf{$\sqrt{\epsilon_{PEHE}}$}   & \multicolumn{3}{c}{\textbf{CEMNIST}($r_{\text{rej}}=0.5$)} \\
                    \textbf{Method / \emph{Pol.}}       & \textbf{\emph{rand.}} & \textbf{\emph{prop.}} & \textbf{\emph{unct.}} \\ \midrule
                    \textbf{BART}                       & 2.1$\pm$.0\          & 2.1$\pm$.0\          & \textbf{2.0$\pm$.0}\ \\
                    \textbf{BT-Learner}                 & 0.3$\pm$.0\          & 0.2$\pm$.0\          & \textbf{0.0$\pm$.0}\ \\
                    \textbf{BTARNet}                    & 0.2$\pm$.0\          & 0.2$\pm$.0\          & \textbf{0.0$\pm$.0}\ \\
                    \textbf{BCFR-MMD}                   & 0.3$\pm$.0\          & 0.3$\pm$.0\          & \textbf{0.1$\pm$.0}\ \\
                    \textbf{BDragonnet}                 & 0.2$\pm$.0\          & 0.2$\pm$.0\          & \textbf{0.0$\pm$.0}\ \\ 
                    \textbf{BCEVAE}                     & 0.3$\pm$.0\          & 0.2$\pm$.0\          & \textbf{0.0$\pm$.0}\ \\
                \bottomrule
                \end{tabular}
            \end{small}
        }
    \end{subfigure}
    
    \caption{\textbf{CEMNIST} evaluation. (a) Histogram of estimated propensity scores. Untreated nines account for the peaks on the left side. (b) Error rate for different rejection policies as we vary the rejection rate. (c) $\sqrt{\epsilon_{PEHE}}$ for different models at a fixed rejection rate $r_{\text{rej}}=0.5$. Compared are the policies \textit{random}, \emph{propensity trimming}, and  \emph{epistemic uncertainty}.
    }
    \label{fig:cemnist_vis}
\end{figure*}

This experimental setup is chosen to demonstrate where the \emph{propensity} based rejection policies can be inappropriate for the prediction of individual-level causal effects. 
Figure \ref{fig:cemnist_vis_hist} shows the histogram over training set predictions for a deep propensity model on a realization of the CEMNIST dataset. 
A data scientist following the trimming paradigm \citep{caliendo2008some} would be justified in choosing a lower threshold around 0.05 and an upper threshold around 0.75. 
The upper threshold would properly reject twos, but the lower threshold would start rejecting nines, which represent the population that the CATE estimator can be most confident about. 
Therefore, rejection choices can be worse than random.

Figure \ref{fig:cemnist_vis_error} shows that the recommendation-error-rate is significantly lower for the \emph{epistemic uncertainty} policy (green dash-dot) than for both the  \emph{random} baseline policy (red dot) and the \emph{propensity} based policies (orange dash and blue solid). 
BT-Learner is used for this plot. 
These results hold across a range of other SOTA CATE estimators for the $\sqrt{\epsilon_{PEHE}}$, as shown in figure \ref{fig:cemnist_table}, and in Appendix \ref{app:results:negative_sampling}. 
Details on the protocol generating these results are in Appendix \ref{app:datasets:cemnist}.

\begin{table}[h!] \centering
    \vspace{-5mm}
    \caption{Comparing \emph{epistemic uncertainty}, \emph{propensity trimming}, and \emph{random} rejection policies for IHDP, IHDP Covariate Shift, and ACIC 2016 and with uncertainty-equipped SOTA models. 50\% or 10\% of examples set to be rejected and errors are reported on the remaining test-set recommendations. \emph{Epistemic uncertainty} policy leads to the lowest errors in CATE estimates (in bold).}
    \vspace{0.1in}
    \begin{small}
        \begin{tabular}{@{}lrrr|rrr|rrr@{}}\toprule
            \textbf{$\sqrt{\epsilon_{PEHE}}$} & \multicolumn{3}{c|}{\textbf{IHDP} ($r_{\text{rej}}=0.1$)} & \multicolumn{3}{c|}{\textbf{IHDP Cov. ($r_{\text{rej}}=0.5$)}} & \multicolumn{3}{c}{\textbf{ACIC 2016} ($r_{\text{rej}}=0.1$)} \\
            \textbf{Method / \emph{Pol.}} & \textbf{\emph{rand.}} & \textbf{\emph{prop.}} & \textbf{\emph{unct.}} & \textbf{\emph{rand.}} & \textbf{\emph{prop.}} & \textbf{\emph{unct.}} & \textbf{\emph{rand.}} & \textbf{\emph{prop.}} & \textbf{\emph{unct.}} \\ \midrule
            \textbf{BART} & 1.9$\pm$.2\ & 1.9$\pm$.2\ & \textbf{1.6$\pm$.1}\ & 2.6$\pm$.2\ & 2.7$\pm$.3\ & \textbf{1.8$\pm$.2}\ & 1.3$\pm$.1\ & 1.2$\pm$.1\ & \textbf{0.9$\pm$.1}\  \\ 
            \textbf{BT-Learner} & 1.0$\pm$.0\ & 0.9$\pm$.0\ & \textbf{0.7$\pm$.0}\ & 2.3$\pm$.2\ & 2.3$\pm$.2\ & \textbf{1.3$\pm$.1}\ & 2.1$\pm$.1\ & 2.0$\pm$.1\ & \textbf{1.5$\pm$.1}\  \\ 
            \textbf{BTARNet}    & 1.1$\pm$.0\ & 1.0$\pm$.0\ & \textbf{0.8$\pm$.0}\ & 2.2$\pm$.3\ & 2.0$\pm$.3\ & \textbf{1.2$\pm$.1}\ & 1.8$\pm$.1\ & 1.7$\pm$.1\ & \textbf{1.2$\pm$.1}\  \\ 
            \textbf{BCFR-MMD}   & 1.3$\pm$.1\ & 1.3$\pm$.1\ & \textbf{0.9$\pm$.0}\ & 2.5$\pm$.2\ & 2.4$\pm$.3\ & \textbf{1.7$\pm$.2}\ & 2.3$\pm$.2\ & 2.1$\pm$.1\ & \textbf{1.7$\pm$.1}\  \\ 
            \textbf{BDragonnet} & 1.5$\pm$.1\ & 1.4$\pm$.1\ & \textbf{1.1$\pm$.0}\ & 2.4$\pm$.3\ & 2.2$\pm$.3\ & \textbf{1.3$\pm$.2}\ & 1.9$\pm$.1\ & 1.8$\pm$.1\ & \textbf{1.3$\pm$.1}\  \\ 
            \textbf{BCEVAE}     & 1.8$\pm$.1\ & 1.9$\pm$.1\ & \textbf{1.5$\pm$.1}\ & 2.5$\pm$.2\ & 2.4$\pm$.3\ & \textbf{1.7$\pm$.1}\ & 3.3$\pm$.2\ & 3.2$\pm$.2\ & \textbf{2.9$\pm$.1}\  \\ \bottomrule
        \end{tabular}
    \end{small}\vspace{-2mm}
    \label{tab:cemnist_rejection_overlap}
\end{table}

\subsection{Uncertainty under covariate shift}
\label{sec:experiment_cov_shift}

\textbf{Infant Health and Development Program (IHDP).} When deploying a machine learning system, we must often deal with a test distribution of $\x$ which is different from the training distribution $p(\x)$. 
We induce a covariate shift in the semi-synthetic dataset IHDP \citep{hill2011bayesian, shalit2017estimating} by excluding instances \textit{from the training} set for which the mother is unmarried. 
Mother's marital status is chosen because it has a balanced frequency of $0.52\pm0.00$; furthermore, it has a mild association with the treatment as indicated by a log odds ratio of $2.22\pm0.01$; and most importantly, there is evidence of a simple distribution shift, indicated by a predictive accuracy of $0.75\pm0.00$ for marital status using a logistic regression model over the remaining covariates. 
We comment on the ethical implications of this experimental set-up, describe IHDP, and explain the experimental protocol in Appendix \ref{app:datasets:ihdp}.

\begin{figure*}[h!]
    \centering
    \begin{subfigure}[t]{0.333\textwidth}
        \centering
        \includegraphics[width=\textwidth]{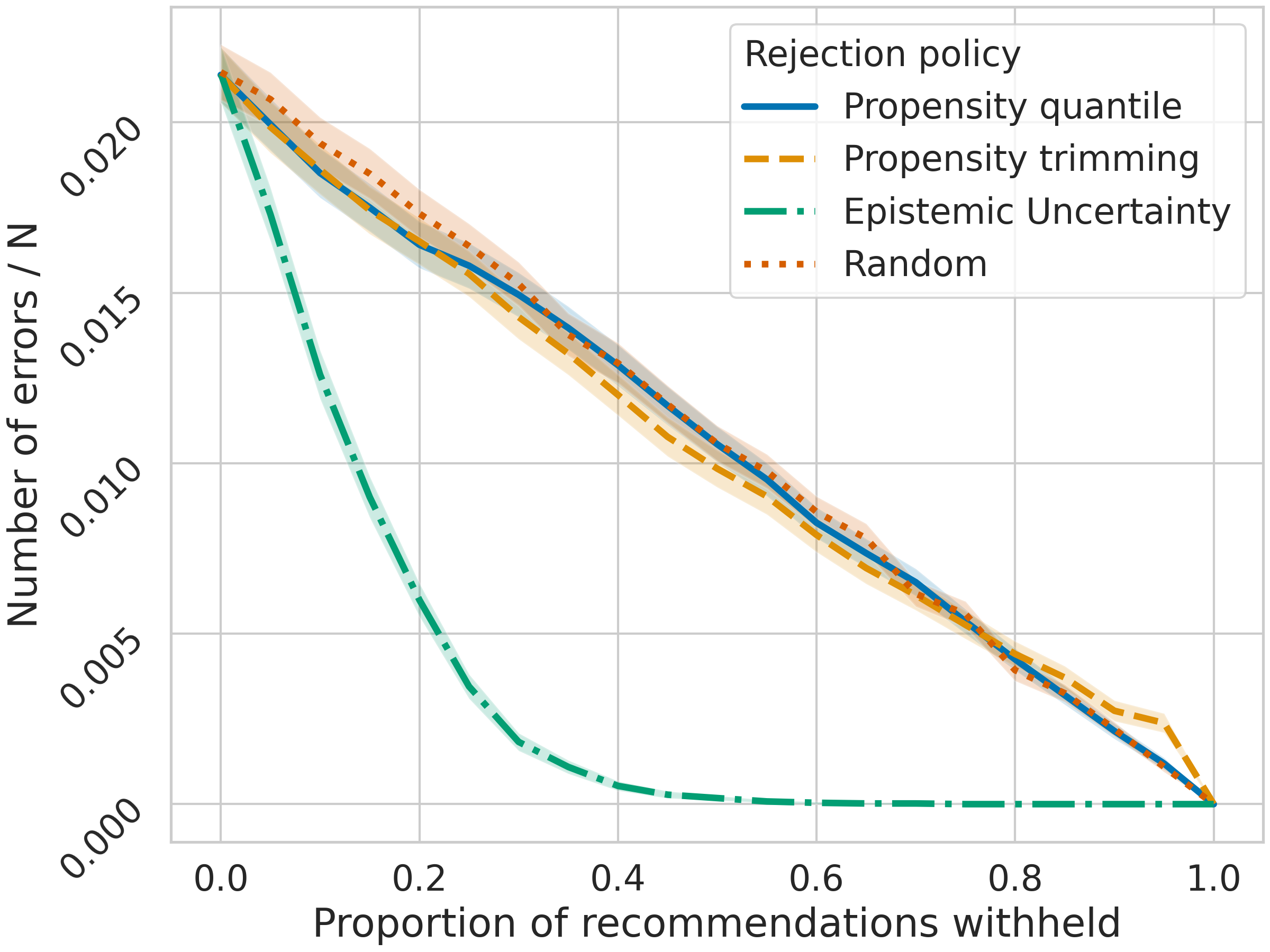}
        \caption{IHDP}
        \label{fig:ihdp_vis_error}
    \end{subfigure}%
    ~ \hspace{-2mm}
    \begin{subfigure}[t]{0.333\textwidth}
        \centering
        \includegraphics[width=\textwidth]{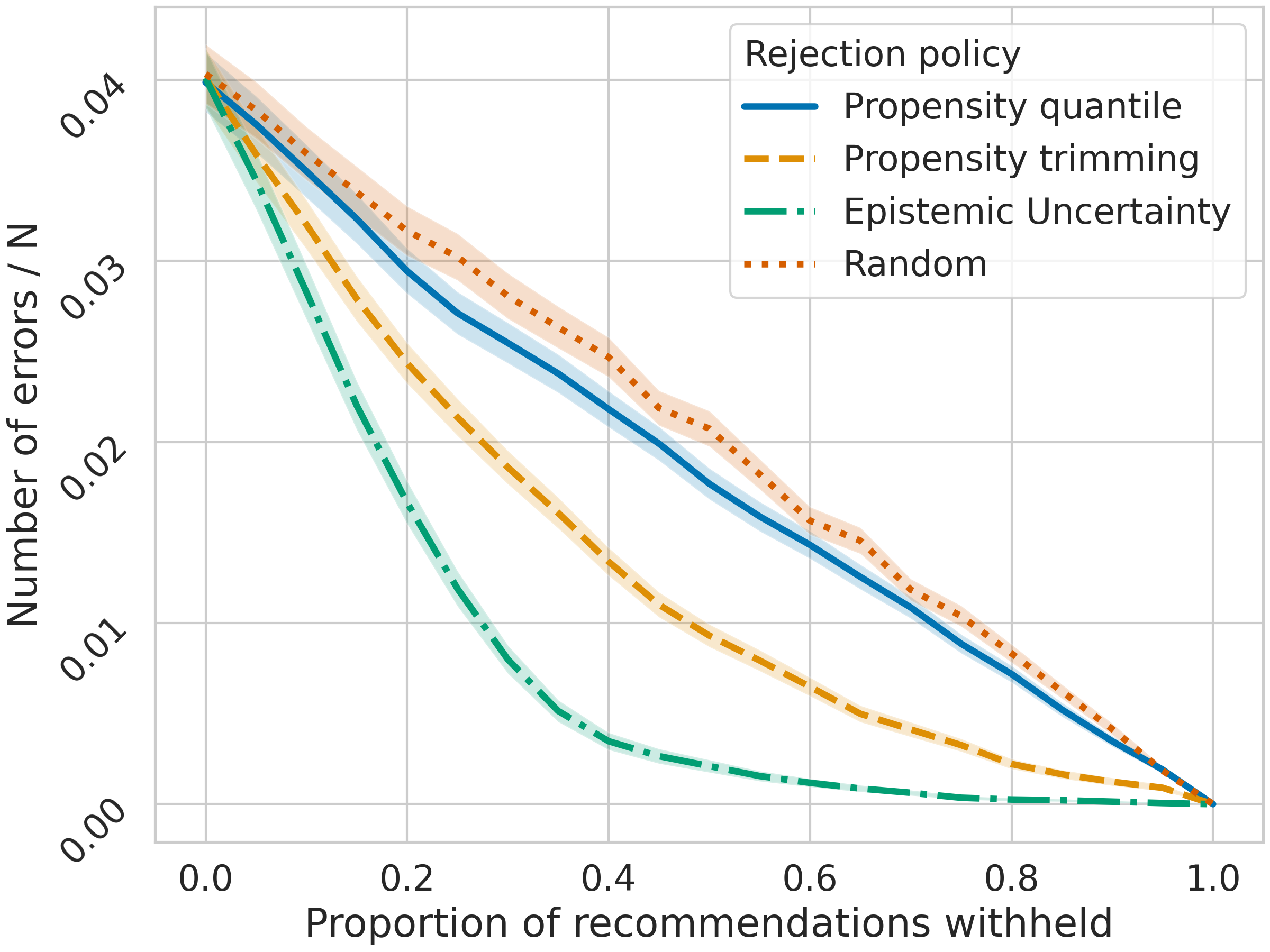}
        \caption{IHDP Covariate Shift}
        \label{fig:ihdp_cov_shift_error}
    \end{subfigure}%
    ~ \hspace{-2mm}
    \begin{subfigure}[t]{0.333\textwidth}
        \centering
        \includegraphics[width=\textwidth]{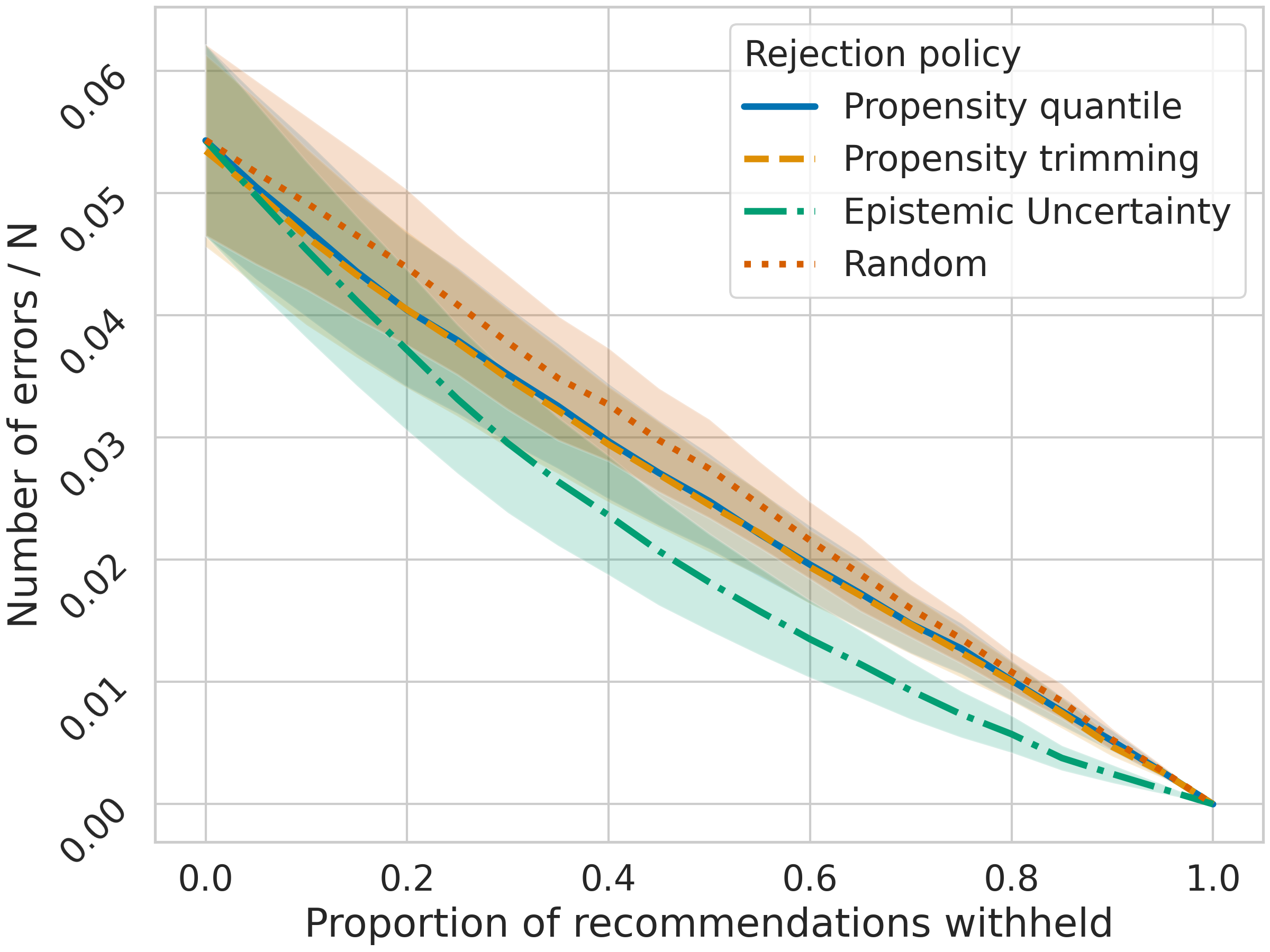}
        \caption{ACIC 2016}
        \label{fig:acic_error}
    \end{subfigure}
    
    \caption{Uncertainty based rejection policies yield significantly lower error rates while withholding fewer recommendations than propensity policies, on IHDP, IHDP Cov., and ACIC 2016.
    } \vspace{-3mm}
    \label{fig:ihdp-cov-vis}
\end{figure*}

We report the mean and standard error in recommendation-error-rates and $\sqrt{\epsilon_{PEHE}}$ over 1000 realizations of the IHDP Covariate-Shift dataset to evaluate each policy by computing each metric over the test set (both sub-populations included). 
We sweep $r_\text{rej}$ from 0.0 to 1.0 in increments of 0.05. Figure \ref{fig:ihdp_cov_shift_error} shows, for the BT-Learner, that the \emph{epistemic uncertainty} (green dash-dot) policy significantly outperforms the uncertainty-oblivious policies across the whole range of rejection rates, and we show in Appendix \ref{app:results} that this trend holds across all models. 
The middle section of table \ref{tab:cemnist_rejection_overlap} supports this claim by reporting the $\sqrt{\epsilon_{PEHE}}$ for each model at $r_\text{rej}=0.5$; the approximate frequency of the excluded population. 
Every model class shows improved rejection performance. 
However, comparisons between model classes are not necessarily appropriate since some models target different scenarios, for example, CEVAE targets \textit{non}-synthetic data where confounders $\z$ aren't directly observed, and it is known to underperform on IHDP \citep{cevae_louizos2017causal}.

We report results for the unaltered IHDP dataset in figure \ref{fig:ihdp_vis_error} and the l.h.s. of table \ref{tab:cemnist_rejection_overlap}. 
This supports that uncertainty rejection is more \emph{data-efficient}, i.e., errors are lower while rejecting less. 
This is further supported by the results on ACIC 2016 \citep{dorie2019automated} (figure \ref{fig:acic_error} and the r.h.s. of table \ref{tab:cemnist_rejection_overlap}). 
The preceding results can be reproduced using publicly available code\footnote{Available at: \url{https://github.com/OATML/ucate}}.

\section{Conclusions}
\label{sec:conclusions}

Observational data often violates the crucial overlap assumption, especially when the data is high-dimensional \citep{d2017overlap}. 
When these violations occur, causal inference can be difficult or impossible, and ideally, a good causal model should communicate this failure to the user. 
However, the only current approach for identifying these failures in deep models is via the propensity score. 
We develop here a principled approach to modeling outcome uncertainty in individual-level causal effect estimates, leading to more accurate identification of cases where we cannot expect accurate predictions, while the propensity score approach can be both over- and under-conservative. 
We further show that the same uncertainty modeling approach we developed can be usefully applied to predicting causal effects under covariate shift. 
More generally, since causal inference is often needed in high-stakes domains such as medicine, we believe it is crucial to effectively communicate uncertainty and refrain from providing ill-conceived predictions.

\section{Broader impact}
Here, we highlight a set of beneficial and potentially alarming application scenarios. 
We are excited about our methods to contribute to ongoing efforts to create neural treatment recommendation systems that can be safely used in medical settings. 
Safety, along with performance, is a major roadblock for this application. 
In regions where medical care is scarce, it may be especially likely that systems will be deployed despite limited safety, leading to potentially harmful recommendations. 
In regions with more universal medical care, individual-based recommendations could improve health outcomes, but systems are unlikely to be deployed when they are not deemed safe.

Massive observational datasets are available to consumer-facing online businesses such as social networks, and to some governments. 
For example, standard inference approaches are limited for recommendation systems on social media sites because a user's decision to follow a recommendation (the treatment) is confounded by the user's attributes (and even the user-base itself can be biased by the recommendation algorithm's choices) \citep{schnabel2016recommendations}. 
Causal approaches are therefore advantageous. 
Observational datasets are typically high-dimensional, and therefore likely to suffer from severe overlap violations, making the data unusable for causal inference, or implying the need for cumbersome preprocessing. 
As our methods enable working directly with such data, they might enable the owners of these datasets to construct causal models of \textit{individual} human behavior, and use these to manipulate attitudes and behavior. 
Examples of such manipulation include affecting voting and purchasing choices.

\section{Acknowledgements}
We would like to thank Lisa Schut, Clare Lyle, and all anonymous reviewers for their time, effort, and valuable feedback. 
S.M. is funded by the Oxford-DeepMind Graduate Scholarship. 
U.S. was partially supported by the Israel Science Foundation (grant No. 1950/19).

\bibliography{references}
\bibliographystyle{icml2020} 

\newpage
\begin{appendices}

\section{Datasets}
\label{app:datasets}

\subsection{CEMNIST}
\label{app:datasets:cemnist}

\begin{table}[h!] \centering
    \caption{\textbf{CEMNIST-Overlap} Description of ``Causal effect MNIST'' dataset.}
    \vspace{0.1in}
    \begin{small}
        \begin{tabular}{@{}lccccc@{}}\toprule
            \textbf{Digit(s)} & Number of train samples & Number treated & $Y^0$ & $Y^1$ & $\CATE$   \\ \midrule
            $9$ & $6000$\         & $\approx 666$\ & $1$\  & $0$\  & $-1$\  \\ 
            $2$ & $\approx 666$ \ & $\approx 666$\          & $0$\  & $1$\  & $1$\   \\ 
            other odds & $\approx 666$ each\ & $\approx 333$ each\        & $1$\  & $0$\  & $-1$\  \\ 
            other evens & $\approx 666$ each\ & $\approx 333$ each\        & $0$\  & $1$\  & $1$\  \\
            \bottomrule
        \end{tabular}
    \end{small}
    \label{tab:cemnist-overlap-appendix}
\end{table}

The original MNIST image dataset contains a training set of size 60000 and a test set of size 10000, where each digit class 0-9 represents 10\% of points. We use a subset of the training data, shown in Table \ref{tab:cemnist-overlap-appendix}. Similarly, we use a subset of the test set, with the same proportion for each digit as in the training set (and the same proportion of treated points). The variables $Y^1$, $Y^0$ are deterministic as shown in Table \ref{tab:cemnist-overlap-appendix}. Some numbers in Table \ref{tab:cemnist-overlap-appendix} are approximate because they are generated according to the probabilities in Table \ref{tab:cemnist-overlap}.

The dataset serves two purposes. First, it illustrates why the standard practice of rejecting points with propensity scores close to $0$ or $1$ can be worse than rejecting randomly. The digit $9$ has the most data making it easy to predict the $\CATE$, but its propensity score is only $0.1$, so that $9$s will be rejected early. It might be a common situation in practice that a sub-population represents the majority of the data and therefore its $\CATE$ is easy to estimate. Second, the digit $2$ suffers from strict non-overlap (propensity score of $1$). It should be the first digit class to be rejected by any method since its $\CATE$ cannot be estimated. When increasing the rejected proportion, digits other than $9$ should subsequently be rejected as only $334$ and $333$ examples are observed for their treatment and control groups respectively. However, propensity-based rejection is likely to retain these sub-populations because their propensity score is $0.5$.

We repeated the CEMNIST experiment $20$ times, each time generating a new dataset with a different random initialization for each model. Note that this is a single dataset, unlike other causal inference benchmarks, so it is only suited for $\CATE$ estimation, not $\ATE$ estimation.

\subsection{IHDP}
\label{app:datasets:ihdp}
\citet{hill2011bayesian} introduced a causal inference dataset based on the The Infant Health Development Program (IHDP), a randomized experiment that assessed the impact of specialist home visits on children's performance in cognitive tests. Real covariates and treatments related to each participant are used in the IHDP dataset. However, outcomes are simulated based on covariates and treatment, making this dataset semi-synthetic. Covariates were made different between the treatment and control groups by removing units with non-white mothers from the treated population. There are 747 units in the dataset (139 treated, 608 control), with 25 covariates related to the children and their mothers. Following \citet{shalit2017estimating,hill2011bayesian}, we use the simulated outcome implemented as setting “A” in the NPCI package \citep{dorie2016npci} and we use the noiseless/expected outcome to compute the ground truth $\CATE$. The IHDP dataset is available for download at \url{https://www.fredjo.com/}.

We run the experiment according to the protocol described in \citep{shalit2017estimating}: we run 1000 repetitions of the experiment, where each test set has 75 points and the remaining 672 available points are split 70\% to 30\% for training and validation. The ground truth outcomes are normalized to a mean of $0$ and standard deviation of $1$ over the training set. For evaluation, each model's predictions are unnormalized to calculate the PEHE.

\textbf{IHDP Covariate Shift. }
As previously mentioned, we selected a variable (marital status of mother) and exclude datapoints where the mother was unmarried from training (while leaving the test set unaltered). We selected this feature for three reasons: it is active in roughly 50\% of data points, the distributions of the remaining covariates were distinct based on a T-SNE visualization \citep{maaten2008visualizing}, and the feature is only marginally correlated with treatment (which ensures that we study the impact of covariate shift, not unobserved confounding). The feature is hidden to the models to make the detection of covariate shift non-trivial, and to induce a more realistic scenario where latent factors are often unaccounted for in observational data.

Marital status may be considered a sensitive socio-economic factor. We do not intend the experiment to be politically insensitive, rather that it emphasizes the problem of demographic exclusion in observational data due to issues such as historical bias, along with the danger of making confident but uninformed predictions when demographic exclusion is latent. Omitting these variables can lead to subpar model performance -- particularly for members of a socio-economic minority.

\subsection{ACIC 2016}
\label{app:datasets:acic}

\citet{dorie2019automated} introduced a dataset named after the 2016 Atlantic Causal Inference Conference (ACIC) where it was used for a competition. ACIC is a collection of semi-synthetic datasets whose covariates are taken from a large study conducted on pregnant women and
their children to identifying causal factors leading to developmental disorders \citep{niswander1972collaborative}. There are 4802 observations and 58 covariates. Outcomes and treatments are simulated, as in IHDP, according to different data-generating process for each dataset. We chose this dataset instead of the 2018 ACIC challenge \citep{shimoni2018benchmarking} because the latter is aimed at only $\ATE$ estimation and the $\CATE$ is equal for each observation in most datasets.

\subsection{Evaluation metrics}
\label{app:evaluation}

We evaluate our methods by considering \emph{treatment recommendations}. A simplified recommendation strategy for an individual-level treatment of a unit with covariates $x_i$ is to recommend $t=1$ if the predicted $\CATE(x_i)$ is positive, and $t=0$ if negative. However, if there is insufficient knowledge about the $\CATE$ an individual, and a high cost associated with making errors, it may be preferable to withhold the recommendation, and e.g. refer the case for further scrutiny. It is therefore important to have an informed \emph{rejection policy} for a treatment assigned based on a given $\CATE$ estimator.

To evaluate a rejection policy for a $\CATE$ estimator we assign a cost of 1 to making incorrect predictions and a cost of 0 for making a correct recommendation. At a fixed number of rejections, the utility of a policy can be defined as the inverse of the total number of erroneous recommendations made, i.e., if a policy can correctly identify the model's mistakes and refer such patients to a human expert then it should have a higher utility.

\textbf{Rejection policies} We introduce two rejection policies based on the epistemic and predictive uncertainty estimates of an uncertainty aware $\CATE$ estimator. Both policies opt to reject if the uncertainty estimate is greater than a threshold that rejects a given proportion of the training data $r_{\text{reject}}$. The training data is used since there may not be a large enough test set in practice. For all policies, we determine thresholds on the training set to simulate a real-world individual-level recommendation scenario. The \textit{epistemic uncertainty} policy uses a sample-based estimator of the uncertainty in $\CATE$ (second \emph{r.h.s.} term in eq. \eqref{eq:variance_in_cate}) given by 
\begin{equation}
    \widehat{Var}_{\textit{epi}}[\muone(\x_i) - \muzero(\x_i)] \coloneqq \frac{1}{M} \sum_{j=1}^{M} \left( \mu^{\hat{\w}^1_j}(\x_i) - \mu^{\hat{\w}^0_j}(\x_i) \right)^2 - \left( \frac{1}{M} \sum_{j=1}^{M} \mu^{\hat{\w}^1_j}(\x_i) - \mu^{\hat{\w}^0_j}(\x_i) \right)^2,
\label{eq:epistemic_variance}
\end{equation}
where $M$ Monte Carlo samples are taken from each of $q(\w_0, \w_1| \mathcal{D})$. Note that, for the T-Learner, this posterior factorizes into two independent distributions $q(\w_0 | \mathcal{D}), q(\w_1 | \mathcal{D})$ because there are separate models for the outcome given treatment and no treatment. Furthermore, other models share parameters for $\muwzero(\cdot),\muwone(\cdot)$ so the individual parameters in $\w_0, \w_1$ may overlap.
The \textit{predictive} uncertainty policy uses an estimator of eq. \eqref{eq:variance_in_cate}, $\widehat{Var}_{\textit{pred}}[Y^1 - Y^0|\x_i]$, which has the same functional form as in eq. \eqref{eq:epistemic_variance}, but instead of being over the difference in expected values $\mu^{\hat{\w}^t_j}(\x_i)$ of the output distribution it is over samples $y^{\hat{\w}^t_j}(\x_i)$ of the output distribution.

We compare the utility of these policies to a random rejection baseline and two policies based on propensity scores. The first propensity policy (\textit{propensity quantiles}) finds a two sided threshold on the distribution of estimated propensity scores such that a proportion $(1. - r_{\text{reject}})$ of the training data is retained. The second policy (\textit{propensity trimming}) implements a  trimming algorithm following the guidelines proposed by \citet{caliendo2008some}. 

\section{Models}
\label{app:models}

We evaluate and compare each rejection policy using several uncertainty-aware $\CATE$ estimators. The estimators are the Bayesian versions of CEVAE \cite{cevae_louizos2017causal}, TARNet, CFR-MMD \citep{shalit2017estimating}, Dragonnet \citep{dragon2019}, and a deep version of the T-Learner \citep{shalit2017estimating}. Each model is augmented by introducing Bayesian parameter uncertainty and by predicting a distribution over model outputs. For image data, two convolutional bottom layers are added to each model.

Each model is augmented with Bayesian parameter uncertainty by adding dropout with a probability of 0.1 after each layer (0.5 for layers before the output layer), and setting weight decay penalties to be inversely proportional to the number of examples in the training dataset. At test time, uncertainty estimates are calculated over 100 MC samples.

For the Bayesian T-Learner we use two BNNs, each having 5 dense, 200 neuron, layers. Dropout is added after each dense layers, followed by ELU activation functions. A linear output layer is added to each network, with a sigmoid activation function if the target is binary. For image data, we add a 2-layer convolutional neural network module, with 32 and 64 filters per layer. Spatial dropout \citep{tompson2015efficient}, and ELU activations follow each convolutional layer, and the output is flattened before being passed to the rest of the network. For image data, the Bayesian CEVAE decoder is modified by using a transposed convolution block for the part of the decoder that models $p(\x | \z)$. For the propensity policies, we use a propensity model that has the same form as a single branch of the Bayesian T-learner. The propensity model's L2 regularization is tuned for calibration as this is important for propensity models. We also experimented with a logistic regression model which performed worse.

Adam optimization \citep{kingma2014adam} is used with a learning rate of 0.001 (On CEMNIST the learning rate for the BCEVAE is reduced to 0.0002), and we train each model for a maximum of 2000 epochs, using early stopping with a patience of 50.

Aside from these changes, model architectures, optimization strategies and loss weighting follow what is reported in their respective papers. More details can be seen in the attached code.

\subsection{Compute infrastructure}
All neural network models were implemented in Tensorflow 2.2 \citep{abadi2016tensorflow}, using Nvidia GPUs. BART was implemented using the dbarts R package, available at \url{https://cran.r-project.org/web/packages/dbarts/index.html}.

\section{Additional Results}
\label{app:results}

\subsection{CEVAE negative sampling}
\label{app:results:negative_sampling}

\begin{figure*}[!ht]
    \centering
    
    \begin{subfigure}[]{0.9\textwidth}
        \centering
        \includegraphics[width=\textwidth]{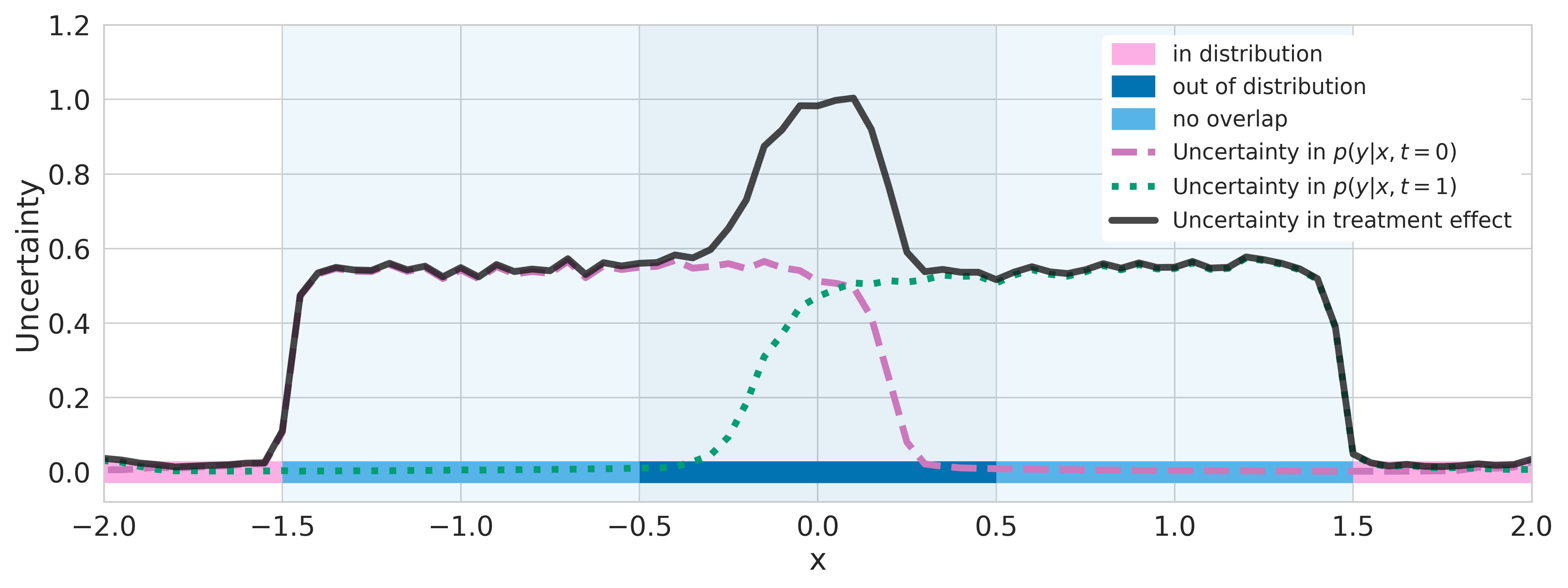}
        \caption{Epistemic uncertainty $\mathcal{I}$ for Bayesian CEVAE trained with negative sampling}
        \label{fig:cevae_uncertainty_ns}
    \end{subfigure}
    ~ 
    
    \begin{subfigure}[]{0.9\textwidth}
        \centering
        \includegraphics[width=\textwidth]{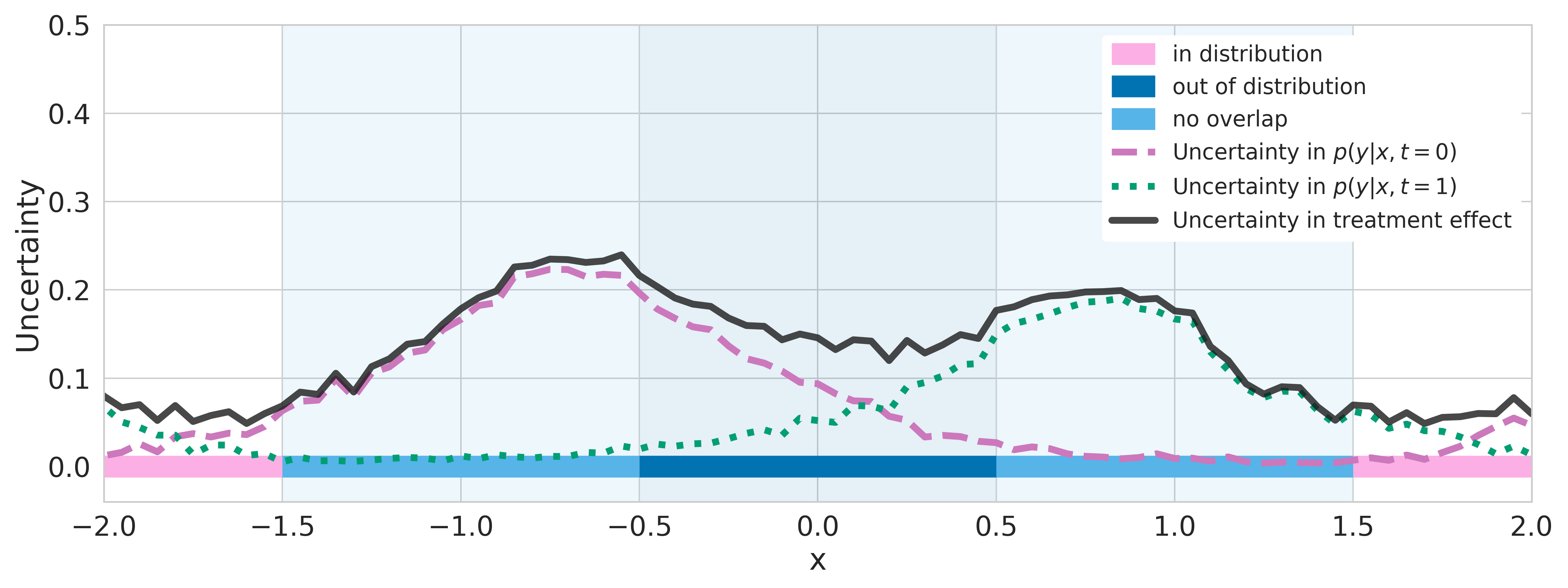}
        \caption{Epistemic uncertainty $\mathcal{I}$ for Bayesian CEVAE trained without negative sampling}
        \label{fig:cevae_uncertainty_nons}
    \end{subfigure}
    \caption{Comparing epistemic uncertainty measures for Bayesian CEVAE trained (a) \emph{with} negative sampling and (b) \emph{without} negative sampling during training. Both models give appropriately low uncertainty measures for in-distribution examples (pink, no shading), and appropriately high uncertainty measures for out-of-distribution examples (dark-blue, dark shading). However, the model trained with negative sampling gives higher uncertainty measures and sharper transitions for non-overlap examples (light-blue, light shading). These properties are important as we propose to use the uncertainty measures to define policies of when to defer a treatment recommendation and instead seek out an expert opinion.
    } \vspace{-6mm}
    \label{fig:cevae_uncertainty_negative_sampling}
\end{figure*}

Figure \ref{fig:cevae_uncertainty_negative_sampling} illustrates the benefit of negative sampling for detecting examples in violation of overlap, or out-of-distribution. Negative sampling results in higher epistemic uncertainty measures and sharper transitions between in-distribution and out-of-distribution regions.

\begin{figure*}[h!]
    \centering
    \begin{subfigure}[t]{0.45\textwidth}
        \centering
        \includegraphics[width=\textwidth]{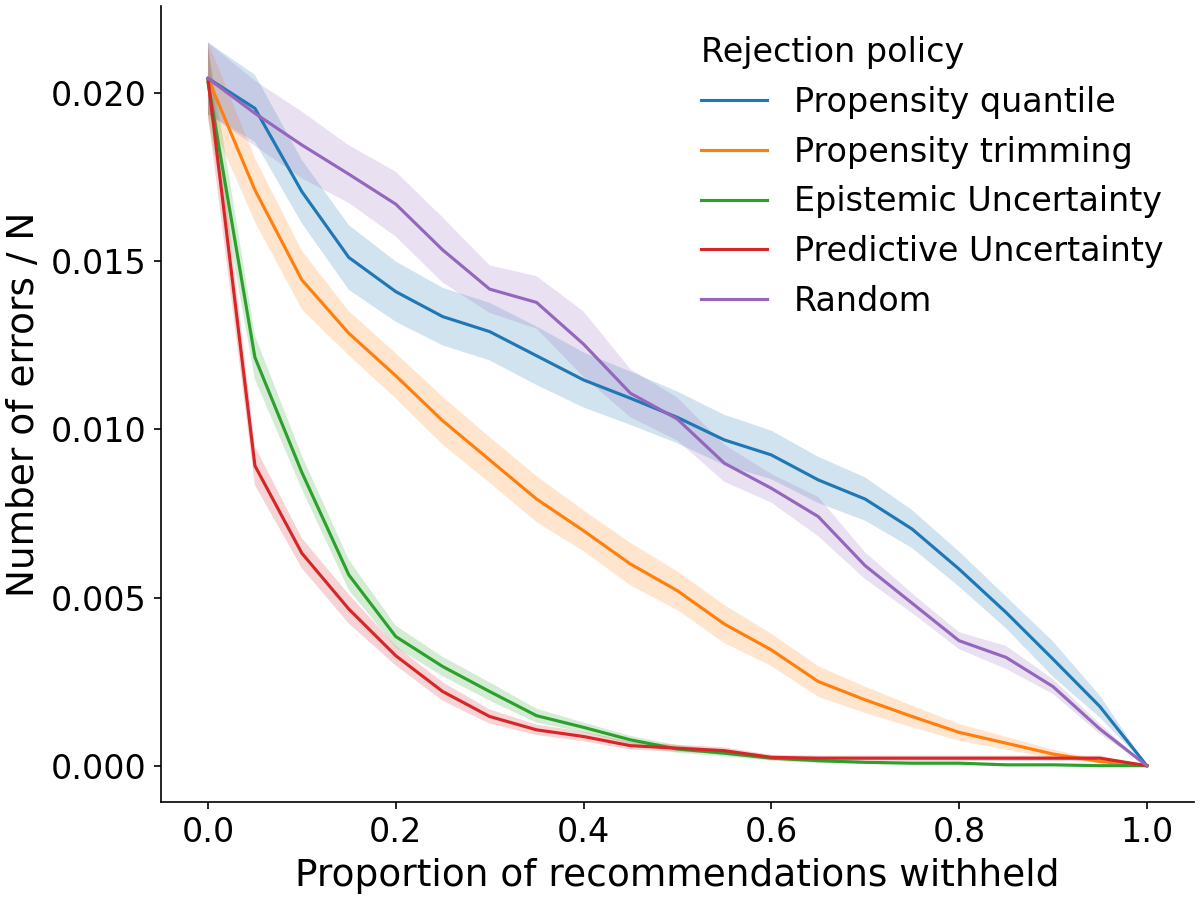}
        \caption{\textbf{Negative Sampling}: Errors}
        \label{fig:error_ns}
    \end{subfigure}%
    ~
    \begin{subfigure}[t]{0.45\textwidth}
        \centering
        \includegraphics[width=\textwidth]{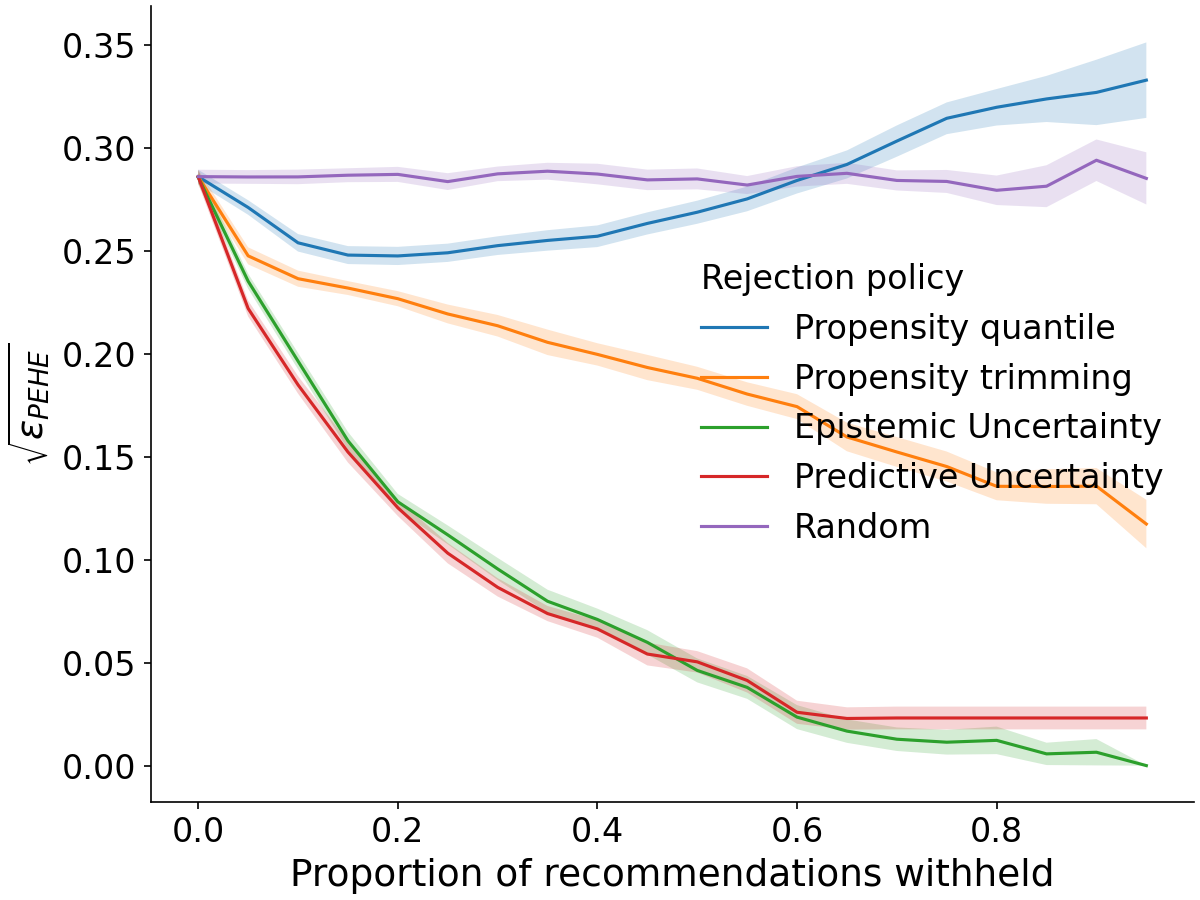}
        \caption{\textbf{Negative Sampling}: $\sqrt{\epsilon_{PEHE}}$}
        \label{fig:pehe_ns}
    \end{subfigure}%
    ~ 
    
    \begin{subfigure}[t]{0.45\textwidth}
        \centering
        \includegraphics[width=\textwidth]{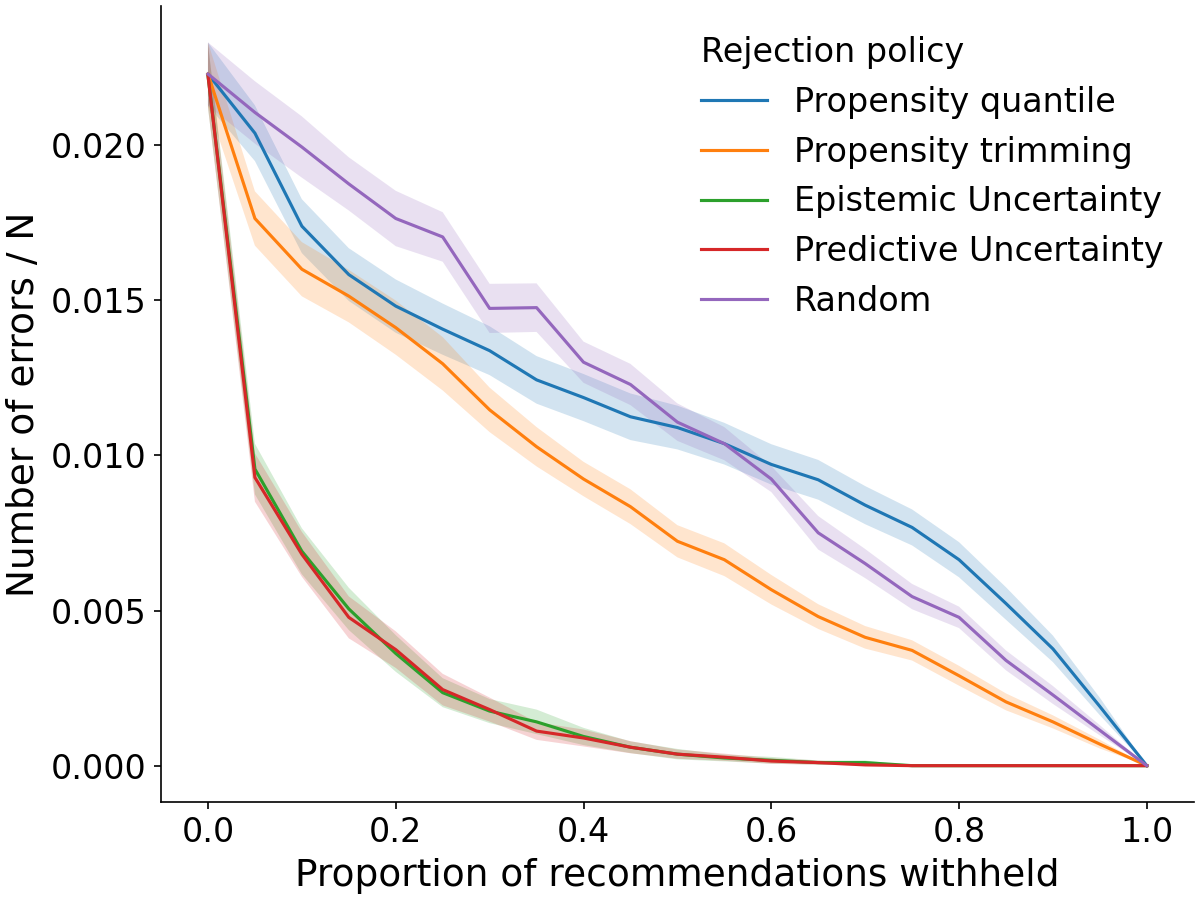}
        \caption{\textbf{No Negative Sampling}: Errors}
        \label{fig:error_nns}
    \end{subfigure}%
    ~
    \begin{subfigure}[t]{0.45\textwidth}
        \centering
        \includegraphics[width=\textwidth]{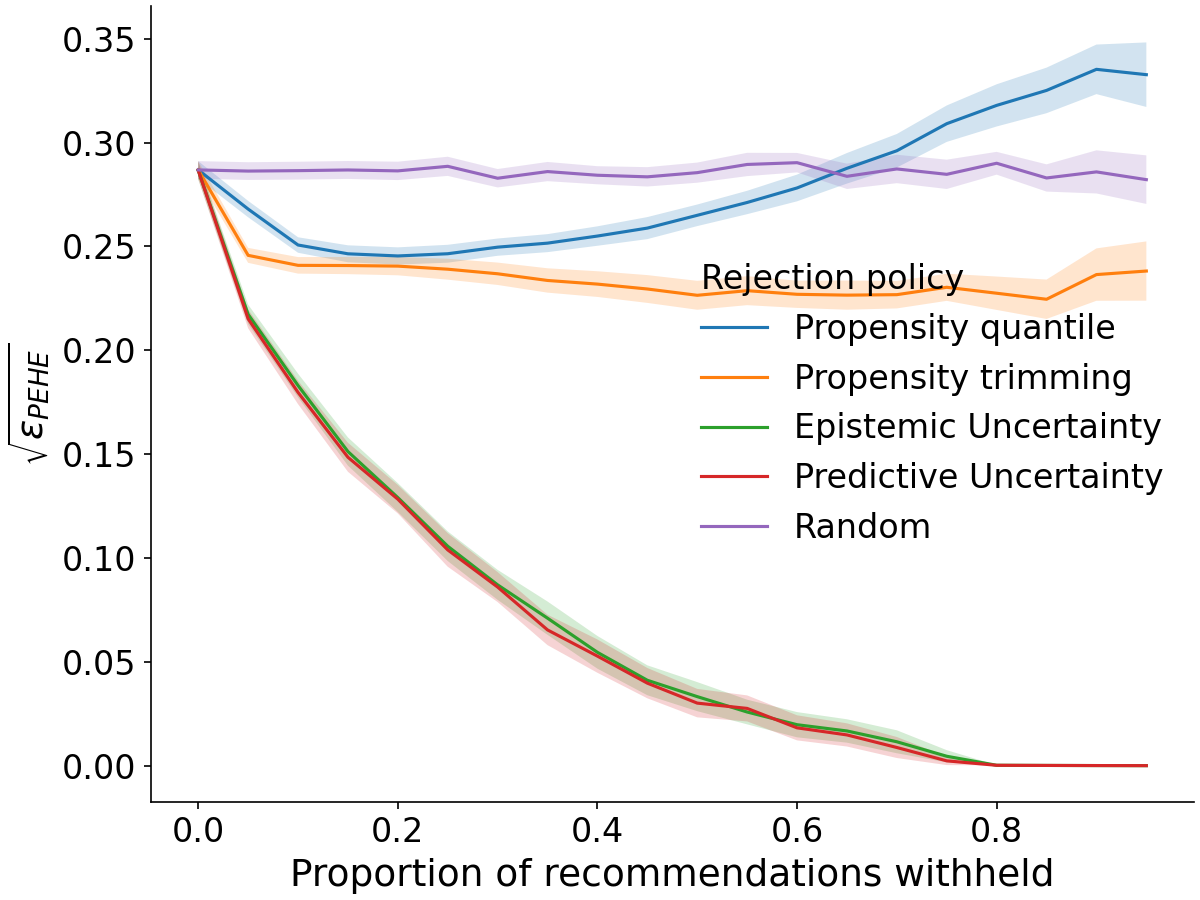}
        \caption{\textbf{No Negative Sampling}: $\sqrt{\epsilon_{PEHE}}$}
        \label{fig:pehe_nns}
    \end{subfigure}%
    ~ 
    \caption{\textbf{CEMNIST BCEVAE} with and without negative sampling.}
    \label{fig:negative_sampling}
\end{figure*}

Table \ref{tab:cemnist_negative_sampling} and figure \ref{fig:negative_sampling} compare the BCEVAE model when trained with and without negative sampling on the CEMNIST dataset.

\begin{table}[h!] \centering
    \caption{Comparing BCEVAE trained with and without-negative sampling on CEMNIST. 50\% of examples set to be rejected and errors are reported on the remaining test-set recommendations. \emph{Epistemic uncertainty} policy leads to the lowest errors in CATE estimates (in bold).}
    \vspace{0.1in}
    \begin{small}
        \begin{tabular}{@{}lrrr|rrr@{}}\toprule
             & \multicolumn{3}{c|}{\textbf{$\sqrt{\epsilon_{PEHE}}$} ($r_{\text{rej}}=0.5$)} & \multicolumn{3}{c}{ \textbf{Rec. Err.} ($r_{\text{rej}}=0.5$)} \\
            \textbf{Method / \emph{Pol.}} & \textbf{\emph{rand.}} & \textbf{\emph{prop.}} & \textbf{\emph{unct.}} & \textbf{\emph{rand.}} & \textbf{\emph{prop.}} & \textbf{\emph{unct.}} \\ \midrule
            \textbf{Negative Sampling} & .295$\pm$.005\ & .227$\pm$.007\ & \textbf{.037$\pm$.009}\ & .010$\pm$.001\ & .005$\pm$.001\ & \textbf{.000$\pm$.000}\  \\ \hline
            \textbf{No Negative Sampling} & .286$\pm$.005\ & .226$\pm$.007\ & \textbf{.033$\pm$.007}\ & .011$\pm$.001\ & .007$\pm$.001\ & \textbf{.000$\pm$.000}\  \\ \bottomrule
        \end{tabular}
    \end{small}
    \label{tab:cemnist_negative_sampling}
\end{table}

\subsection{IHDP}

Table \ref{tab:ihdp} shows the relative performance of the Bayesian models to the results reported in their respective papers for the IHDP dataset.

\begin{table}[h!] \centering
    \caption{Errors on unaltered IHDP, comparing to previously published results (given in the upper half). Note that BCEVAE (ours) outperforms CEVAE. $\epsilon_{ATE}$ is the squared error of the Average Treatment Effect. These results are only for completeness and do not contain the evidence for our main findings.}
    \vspace{0.1in}
    \begin{small}
        \begin{tabular}{@{}lrrrr@{}}\toprule
            & \multicolumn{2}{c}{\textbf{within-sample}} & \multicolumn{2}{c}{\textbf{out-of-sample}} \\
            \textbf{Method} & \textbf{$\sqrt{\epsilon_{PEHE}}$} & \textbf{$\epsilon_{ATE}$} & \textbf{$\sqrt{\epsilon_{PEHE}}$} & \textbf{$\epsilon_{ATE}$} \\ \midrule
            \textbf{OLS-2} & 2.4$\pm$.1\ & .14$\pm$.01\ & 2.5$\pm$.1\ & .31$\pm$.02\ \\ 
            \textbf{BART} & 2.1$\pm$.1\ & .23$\pm$.01\ & 2.3$\pm$.1\ & .34$\pm$.02\ \\ 
            \textbf{BNN} & 2.2$\pm$.1\ & .37$\pm$.03\ & 2.1$\pm$.1\ & .42$\pm$.03\ \\ 
            \textbf{GANITE} & 1.9$\pm$.4\ & .43$\pm$.05\ & 2.4$\pm$.4\ & .49$\pm$.05\ \\ 
            \textbf{CEVAE} & 2.7$\pm$.1\ & .34$\pm$.01\ & 2.6$\pm$.1\ & .46$\pm$.02\ \\ 
            \textbf{TARNet} & .88$\pm$.0\ & .26$\pm$.01\ & .95$\pm$.0\ & .28$\pm$.01\ \\ 
            \textbf{CFR-MMD} & .73$\pm$.0\ & .3$\pm$.01\ & .78$\pm$.0\ & .31$\pm$.01\ \\ 
            \textbf{Dragonnet} & \ & .14$\pm$.01\ & \ & .20$\pm$.01\ \\ \midrule 
            \textbf{BT-Learner} & .95$\pm$.0\ & .21$\pm$.01\ & .88$\pm$.0\ & .18$\pm$.01\ \\  
            \textbf{BTARNet} & 1.1$\pm$.0\ & .23$\pm$.01\ & .96$\pm$.0\ & .20$\pm$.01\ \\  
            \textbf{BCFR-MMD} & 1.3$\pm$.1\ & .29$\pm$.01\ & 1.2$\pm$.1\ & .26$\pm$.01\ \\  
            \textbf{BDragonnet} & 1.5$\pm$.1\ & .30$\pm$.01\ & 1.3$\pm$.0\ & .27$\pm$.01\ \\ 
            \textbf{BCEVAE} & 1.8$\pm$.1\ & .47$\pm$.01\ & 1.8$\pm$.1\ & .50$\pm$.02\ \\ 
            \bottomrule
        \end{tabular}
    \end{small}
    \label{tab:ihdp}
\end{table}

\end{appendices}

\end{document}